\documentclass[final]{cvpr}

\usepackage{times}
\usepackage{epsfig}
\usepackage{graphicx}
\usepackage{amsmath}
\usepackage{amssymb}

\usepackage{subcaption}
\usepackage{multirow}
\usepackage{multicol}
\usepackage{array}
\usepackage[dvipsnames]{xcolor}
\usepackage{bm}

\usepackage[pagebackref=true,breaklinks=true,colorlinks,bookmarks=false]{hyperref}

\def\cvprPaperID{14} 
\def\confYear{CVPR 2021}
\newcommand{\mr}[1]{\multirow{2}{*}{#1}}
\newcommand{\PreserveBackslash}[1]{\let\temp=\\#1\let\\=\temp}
\newcolumntype{C}[1]{>{\PreserveBackslash\centering}p{#1}}
\newcolumntype{R}[1]{>{\PreserveBackslash\raggedleft}p{#1}}
\newcolumntype{L}[1]{>{\PreserveBackslash\raggedright}p{#1}}

\begin{document}

\title{Is In-Domain Data Really Needed? \\A Pilot Study on Cross-Domain Calibration for Network Quantization}

\author{Haichao Yu$^1$, Linjie Yang$^{2}$, Humphrey Shi$^{1}$\\
$^{1}$University of Illinois at Urbana-Champaign, $^{2}$ByteDance Inc.\\
{\tt\small haichao3@illinois.edu, linjie.yang@bytedance.com, shihonghui3@gmail.com}
}

\maketitle

\begin{abstract}
Post-training quantization methods use a set of calibration data to compute quantization ranges for network parameters and activations. The calibration data usually comes from the training dataset which could be inaccessible due to sensitivity of the data. In this work, we want to study such a problem: \textbf{can we use out-of-domain data to calibrate the trained networks without knowledge of the original dataset?} Specifically, we go beyond the domain of natural images to include drastically different domains such as X-ray images, satellite images and ultrasound images. We find cross-domain calibration leads to surprisingly stable performance of quantized models on 10 tasks in different image domains with 13 different calibration datasets. We also find that the performance of quantized models is correlated with the similarity of the Gram matrices between the source and calibration domains, which can be used as a criterion to choose calibration set for better performance. We believe our research opens the door to borrow cross-domain knowledge for network quantization and compression.
\end{abstract}

\section{Introduction}
With the increasing popularity of deploying neural networks on edge devices, neural network quantization has become a widely studied topic~\cite{courbariaux2016binarized,rastegari2016xnor,liu2018bi,zhu2016trained,li2016ternary,zhou2016dorefa,zhang2018lq,jung2019learning,banner2019post,zhao2019improving}. By quantizing its weights and activations to low-bit integers, a neural network can be stored with smaller size and executed at faster speed with less memory footprint and computational resources.

Most existing network quantization methods can be roughly divided into two groups. One is quantization-aware training (QAT), the other is post-training quantization (PQ). By inserting differentiable simulated quantization operations into the network during training, QAT methods allow training losses to back-propagate through the quantization operations. Although QAT methods can achieve satisfactory performance in most cases, it is time-consuming and data-hungry, which is unacceptable in some scenarios, \eg, when the time budget is restricted or training dataset is inaccessible. Compare with QAT, PQ has advantages on time and data efficiency. Given a pretrained full-precision model, PQ runs the model over a small set of calibration data to calculate value ranges of intermediate activations. These calibration data shares the same distribution as the training data on which the model is trained. After this, both network activations and weights can be quantized into low-precision integers for inference acceleration. Compared with QAT, this post-training quantization process is highly efficient (\eg, in a few minutes). 

\begin{figure}[t]
    \centering
    \includegraphics[scale=0.4]{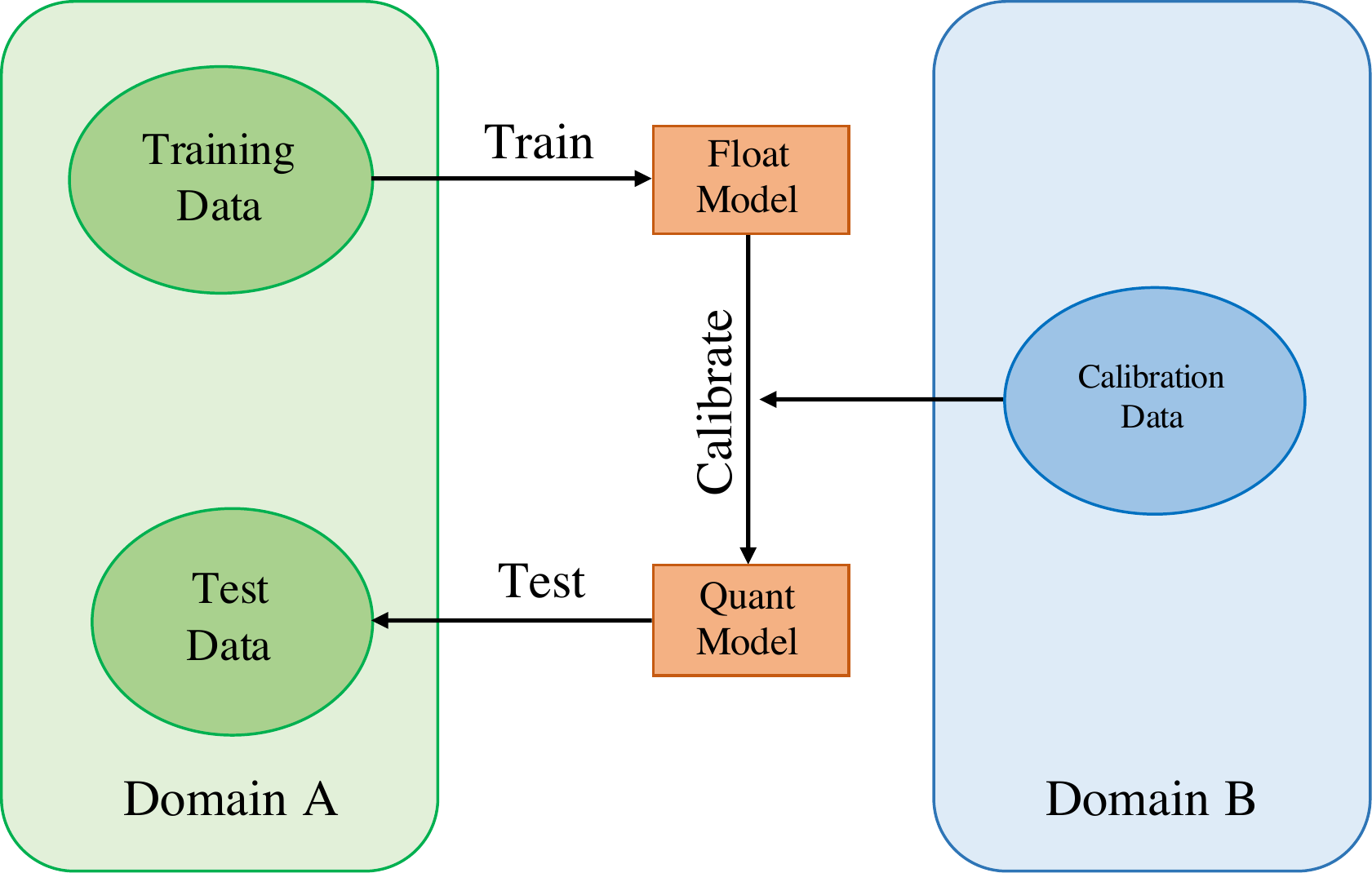}
    \caption{Pipeline of cross-domain calibration for post-training quantization without using in-domain data.}
    \label{fig:teaser}
\end{figure}

However, PQ methods still require in-domain calibration data, which may be inaccessible in some situations due to privacy or security reasons. For example, federated learning~\cite{kon2015federated} trains models using distributed private user data that are not accessible by the developers. Without calibration data from the same domain as the training data, the traditional PQ methods will fail. Targeting at this problem, Cai~\etal proposed a novel method termed ZeroQ that generates synthetic data for calibration~\cite{cai2020zeroq}. Specifically, ZeroQ utilized BatchNorm~\cite{batchnorm-ioffe15} statistics from the pretrained model as supervision to optimize randomly initialized input such that the BatchNorm statistics of the input are similar to those from the pretrained model. ZeroQ showed its effectiveness on various datasets including ImageNet~\cite{deng2009imagenet} and MSCOCO~\cite{lin2014microsoft}. Another method is to use Generative Adversarial Networks (GAN) to create fake calibration data~\cite{xu2020generative}. The proposed framework named GDFQ is trained with BatchNorm statistics loss, Cross Entropy classification loss and knowledge distillation loss.

One drawback of these data-free quantization methods is their high time complexity. With common hardware setting (\eg, on single GPU), ZeroQ requires hundreds of back-propagation iterations to synthesize a single batch and GDFQ needs to be trained for hundreds of epochs. On the other hand, due to the optimization nature of these methods, hyper-parameters also need to be tuned for satisfactory results. If in-domain training data is inaccessible 
and synthetic data generation is time-consuming, a straightforward question to ask is ``Can we use real-world images from another domain for calibration?''. 
As shown in Figure~\ref{fig:teaser}, we use calibration data from domain $B$ to calibration model trained on domain $A$, resulting in a quantized model. The model is further evaluated on a test set in domain $A$. 

To study this problem, we carry out a large-scale benchmark with 10 different tasks and 13 diversified image domains including nature images, low-resolution images, ultrasound images, satellite images, etc. We find that a simple treatment on the BatchNorm layers in the calibration procedure greatly improves performance of cross-domain calibration, almost bridging the gap between in-domain and out-of-domain calibration on a wide range of tasks. 
We also find that the performance of quantized models is correlated with the similarity of the Gram matrices between the source and calibration domains, which can be used as a criterion to choose similar image domains for better performance. Compared with synthetic data generation method ZeroQ, our approach achieves comparable or better performance on our large-scale benchmark. Although the study is mainly empirical, it reveals that cross-domain data can be used for post-training calibration just as effective as in-domain data, which could motivate the community to explore more in the direction of model quantization and compression with cross-domain knowledge.

\section{Related Work}
\subsection{Quantization-Aware Training}
Quantization-aware training (QAT) inserts simulated quantization operation into model forward and backward passes during training. First introduced by Courbariaux~\etal~\cite{hubara2016binarized}, binary neural networks achieved seven times faster inference speed on GPU. Further in \cite{rastegari2016xnor}, Rastegari~\etal proposed to use a binary tensor and a float scalar to approximate the weight or activation tensor. Parallel to binarization, multi-bit model quantization is also widely explored. DoReFa-Net~\cite{zhou2016dorefa} was proposed to quantize both network and training gradients. Scale-adjusted training~\cite{jin2019efficient} improve the model performance by scaling activation in the network to preserve proper scale of gradients. To bring flexibility to the model quantizer, some works also introduced learnable quantization schemes. In \cite{zhang2018lq}, LQ-Net was proposed to learn the quantization basis vectors. In \cite{jung2019learning}, Jung~\etal proposed a novel method to learn the quantization intervals.

\subsection{Post-training Quantization}
Different from QAT, post-training quantization (PQ) directly computes quantization parameters with a calibration set. In \cite{krishnamoorthi2018quantizing}, Krishnamoorthi~\etal introduced PQ with layer-wise and channel-wise quantization schemes. To improve quantization performance, Banner~\etal proposed a novel per-channel PQ scheme that analytically computes clipping thresholds and bits allocated for each channel~\cite{banner2019post}. To mitigate the quantization error from the clipping operation, Zhao~\etal introduced a technique named OCS to split the channels with extreme values~\cite{zhao2019improving}.

Traditionally, PQ method requires a calibration dataset which is often sampled from training data to estimate the clipping ranges of model activations. However, in some scenarios, training data may be inaccessible due to data privacy issues. In this case, quantization could be conducted without a calibration set. Assuming features from a BatchNorm layer follow Gaussian distribution, Nagel~\etal directly used BatchNorm statistics (\ie, running means and variances) to estimate activation ranges~\cite{nagel2019data}. The drawback of this method is the estimated ranges may be inaccurate since the distribution of intermediate layers may be different from Gaussian distribution. DFC and ZeroQ proposed to synthesize calibration data by gradient descent under the supervision of BatchNorm statistics~\cite{haroush2020knowledge,cai2020zeroq}.
 Generative Adversarial Networks (GAN) is also utilized to generate synthetic calibration data~\cite{xu2020generative}, in which the generator generates calibration data and the discriminator is the quantized model. One problem with these optimization-based methods is the high time complexity. Hundreds of iterations or even hundreds of epochs of back-propagation are commonly required. In addition, much human efforts are needed for hyper-parameter tuning such as learning rate and optimizer selection. In contrast to these methods, we find that out-of-domain image datasets can serve as calibration data effectively for a wide range of tasks.

\subsection{Batch Normalization and Domain Knowledge}
To adapt a deep neural network to a different domain, Li~\etal proposed Adaptive BatchNorm to update running statistics in target domain~\cite{li2016revisiting}. The central assumption is that domain specific information is encoded in BatchNorm layers. Sharing a similar spirit with this, Li~\etal showed that style transfer can be conducted by matching the BatchNorm statistics between two images~\cite{li2017demystifying}. 
Inspired by the effect of BatchNorm statistics, we design a BatchNorm updating method on out-of-domain calibration dataset to estimate ranges of activations, which effectively improves the performance of cross-domain calibration even when the training and the calibration domains have huge appearance differences.

\section{Cross-domain Calibration for Post-training Quantization}
\subsection{Motivation}
Given a pretrained full-precision model, post-training quantization (PQ) is widely used for inference acceleration. Most existing PQ methods require a set of calibration data to calculate the quantization parameters for weights and activations. Training data is normally used for calibration. However, in some situations, the training data is not available due to privacy or security issues. To solve this problem, Cai~\etal~\cite{cai2020zeroq} proposed to synthesize calibration data using BatchNorm statistics as supervision. Since different types of real images are vastly available, we would like to explore another option: use out-of-domain real images for calibration. We conduct a large-scale empirical study with drastically different datasets and tasks to investigate this setting. 

\subsection{Quantization Schemes}
In all our experiments, we employ layer-wise uniform quantization for both network weights and activations~\cite{krishnamoorthi2018quantizing}. To quantize a tensor $\bm{x}$ by $k$-bit, the quantized tensor $\bm{x}_q$ is calculated as
\begin{align}
    \bm{x}_{int} &= \mathrm{round}\left(\frac{\bm{x}}{s}\right) + z, \nonumber\\
    \bm{x}_{q} &= \mathrm{clip}\left(\bm{x}_{int}, c_l, c_h\right),
\end{align}
where $s$ and $z$ are scale and zero point parameters, $c_l$ and $c_h$ are lower and upper clipping thresholds. For network weight $\bm{w}$, we use a symmetric min-max quantization scheme, which is defined as
\begin{align}
    s_w &= \frac{\mathrm{max}(\left|w_{min}\right|, \left|w_{max}\right|)}{2^{k-1}}, \nonumber\\
    z_w &= 0, \nonumber\\
    c_l &= -2^{k-1}, \nonumber\\
    c_h &= 2^{k-1} - 1,
    \label{eq:sz_weight}
\end{align}
where $w_{min}$ and $w_{max}$ are minimum and maximum of $\bm{w}$. To quantize activation $\bm{a}$, we employ an affine histogram quantization scheme, which is also used in PyTorch\footnote{\url{https://pytorch.org/docs/master/torch.quantization.html}}. To determine the quantization parameters, we have
\begin{align}
    s_a &= \frac{a_h - a_l}{2^k-1}, \nonumber\\
    z_a &= \frac{a_l(2^k-1)}{a_l - a_h}, \nonumber\\
    c_l &= 0, \nonumber \\
    c_h &= 2^{k} - 1,
    \label{eq:sz_act}
\end{align}
where $a_l$ and $a_h$ are lower and upper clipping thresholds for $\bm{a}$. For each layer, a histogram is built to represent the activation distribution. We search for the optimal $a_l$ and $a_h$ values such that the quantization error is minimized with respect to $\bm{a}$.
In all our experiments, we first fold BatchNorm layers to their preceding convolutional or linear layers before computing quantization parameters. 

\subsection{Datasets and Networks}
\begin{figure*}[t]
    \centering
    \begin{subfigure}[t]{0.195\textwidth}
        \centering
        \includegraphics[width=0.99\linewidth]{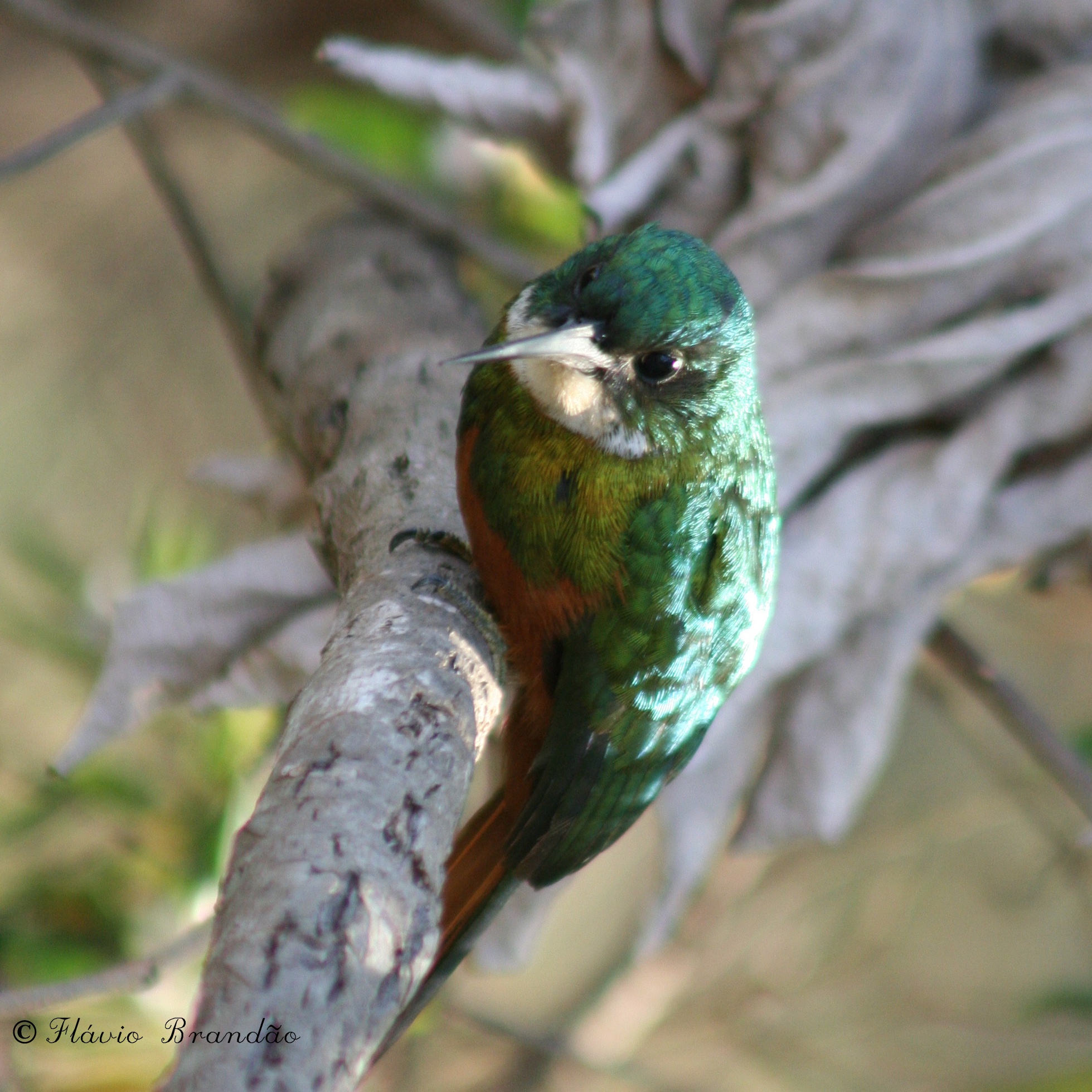}
        \caption{\scriptsize Imagenet~\cite{deng2009imagenet}}
        \label{fig:train_start}
    \end{subfigure}
    \begin{subfigure}[t]{0.195\textwidth}
        \centering
        \includegraphics[width=0.99\linewidth]{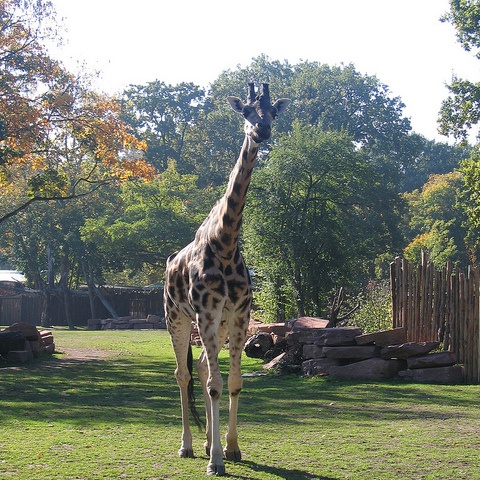}
        \caption{\scriptsize MSCOCO~\cite{lin2014microsoft}}
    \end{subfigure}
    \ 
    \begin{subfigure}[t]{0.39\textwidth}
        \centering
        \includegraphics[width=0.99\linewidth]{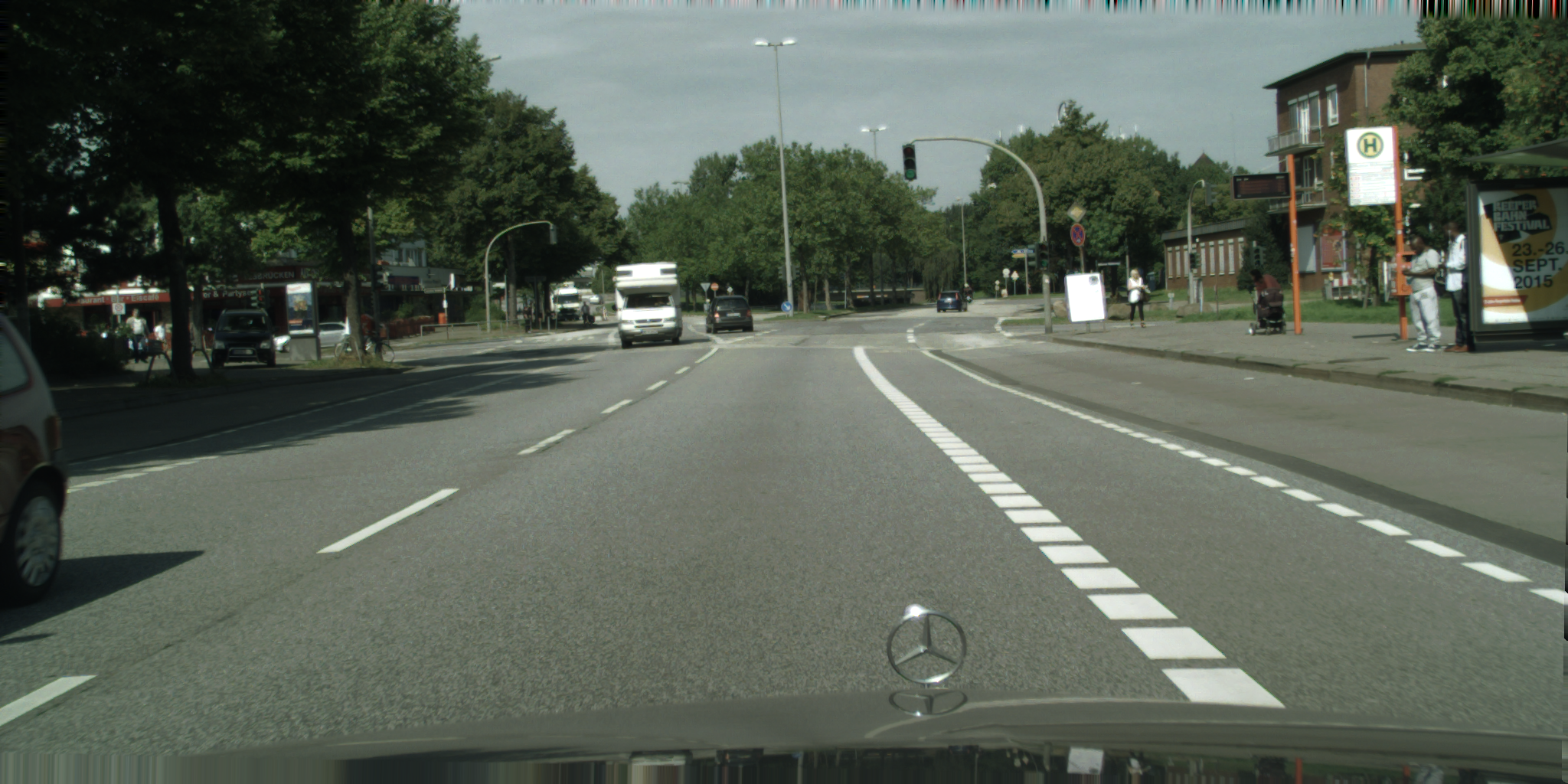}
        \caption{\scriptsize Cityscapes~\cite{cordts2016cityscapes}}
    \end{subfigure}
    \begin{subfigure}[t]{0.195\textwidth}
        \centering
        \includegraphics[width=0.99\linewidth]{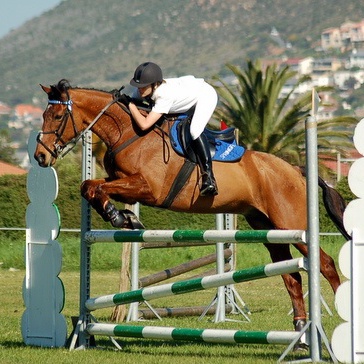}
        \caption{\scriptsize Pascal VOC 07~\cite{pascal-voc-2007}}
    \end{subfigure}
    \begin{subfigure}[t]{0.195\textwidth}
        \centering
        \includegraphics[width=0.99\linewidth]{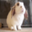}
        \caption{\scriptsize Cifar100~\cite{krizhevsky2009learning}}
    \end{subfigure}
    \begin{subfigure}[t]{0.195\textwidth}
        \centering
        \includegraphics[width=0.99\linewidth]{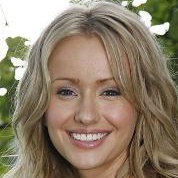}
        \caption{\scriptsize CelebA~\cite{liu2015faceattributes}}
    \end{subfigure}
    \begin{subfigure}[t]{0.195\textwidth}
        \centering
        \includegraphics[width=0.99\linewidth]{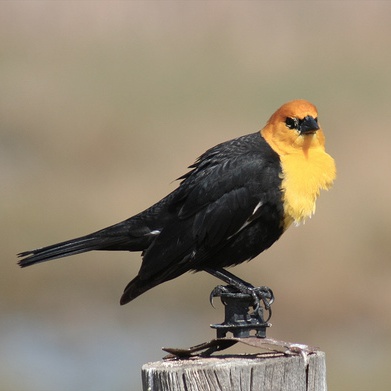}
        \caption{\scriptsize CUB-200-2011~\cite{WahCUB_200_2011}}
    \end{subfigure}
    \ 
    \begin{subfigure}[t]{0.39\textwidth}
        \centering
        \includegraphics[width=0.49\linewidth]{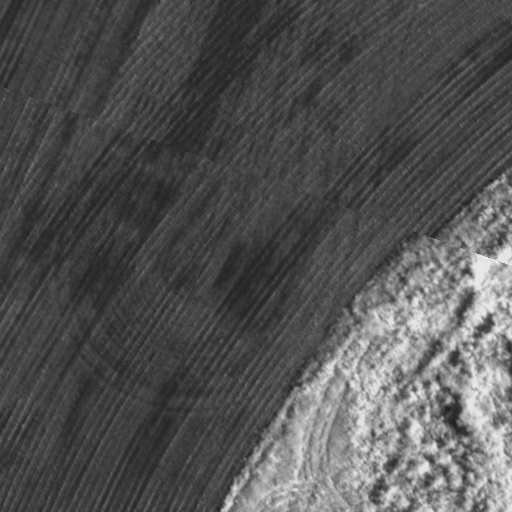}
        \includegraphics[width=0.49\linewidth]{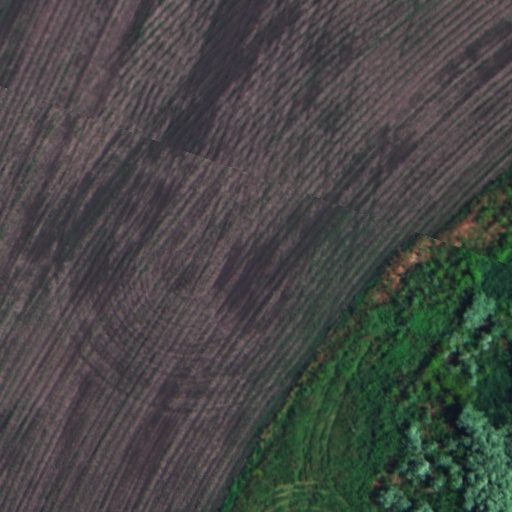}
        \caption{\scriptsize Agriculture (NIR and RGB channel)~\cite{chiu2020agriculture}}
    \end{subfigure}
    \begin{subfigure}[t]{0.195\textwidth}
        \centering
        \includegraphics[width=0.99\linewidth]{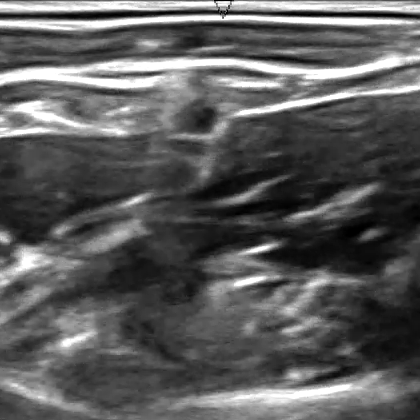}
        \caption{\scriptsize Kaggle Ultrasound~\cite{ultrasound}}
    \end{subfigure}
    \begin{subfigure}[t]{0.195\textwidth}
        \centering
        \includegraphics[width=0.99\linewidth]{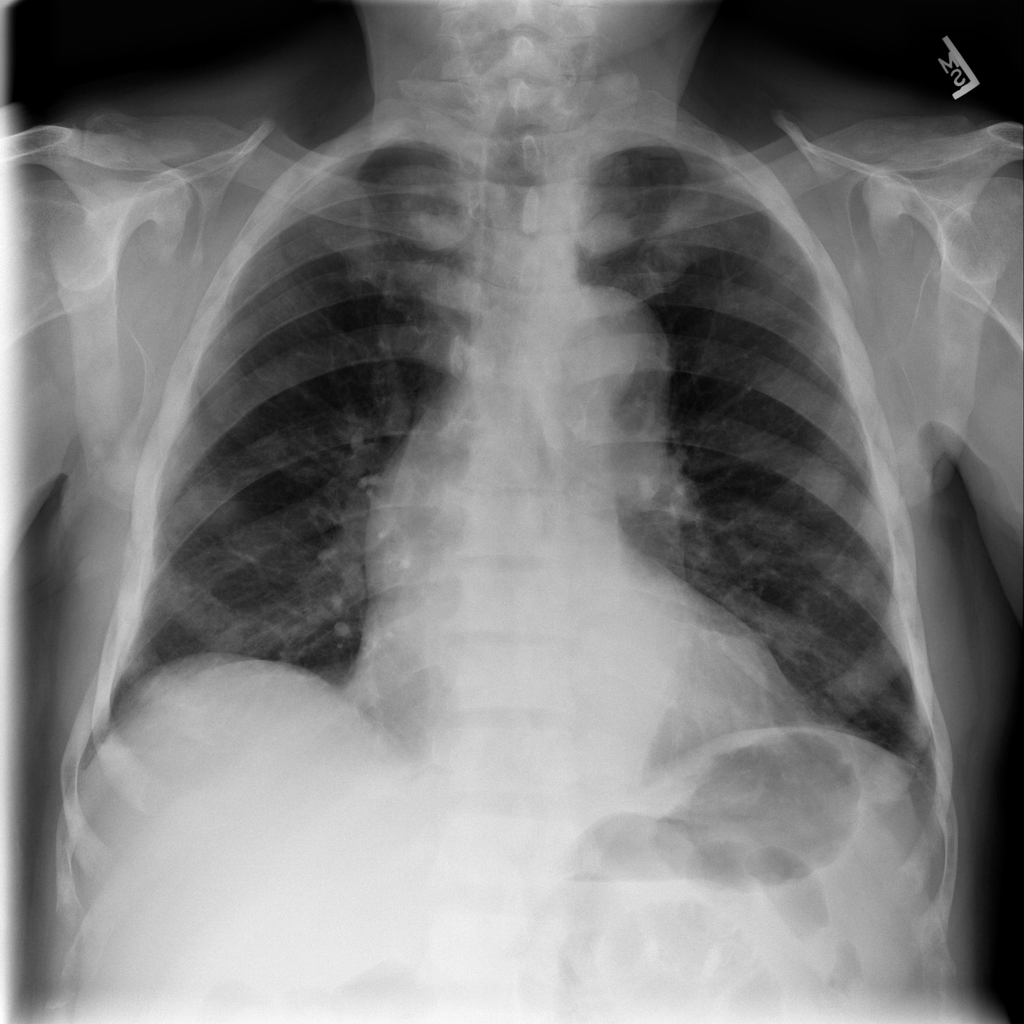}
        \caption{\scriptsize NIH Chest X-ray~\cite{wang2017chestx}}
        \label{fig:train_end}
    \end{subfigure}
    \begin{subfigure}[t]{0.195\textwidth}
        \centering
        \includegraphics[width=0.99\linewidth]{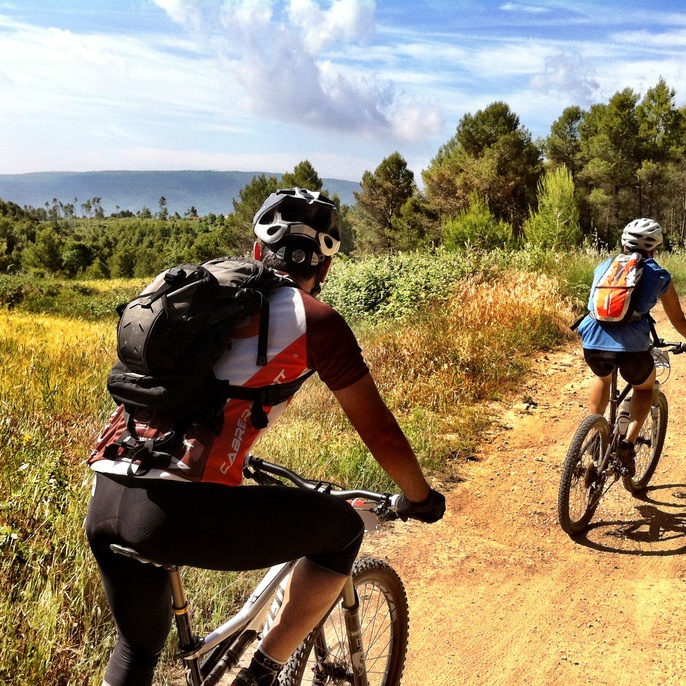}
        \caption{\scriptsize Open Images~\cite{OpenImages}}
        \label{fig:calib_start}
    \end{subfigure}
    \begin{subfigure}[t]{0.195\textwidth}
        \centering
        \includegraphics[width=0.99\linewidth]{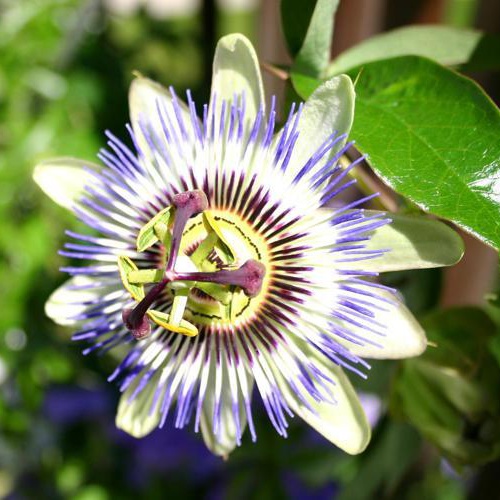}
        \caption{\footnotesize Oxford Flowers~\cite{nilsback2008automated}}
    \end{subfigure}
    \begin{subfigure}[t]{0.195\textwidth}
        \centering
        \includegraphics[width=0.99\linewidth]{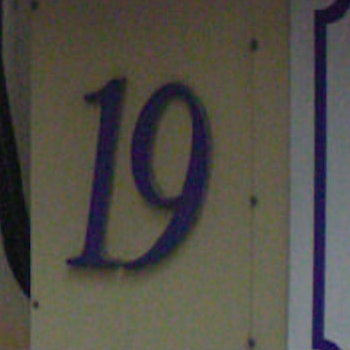}
        \caption{\scriptsize SVHN~\cite{netzer2011reading}}
        \label{fig:calib_end}
    \end{subfigure}
    \caption{Example images from different domains. We will use abbreviations in the following section: IN: ImageNet, CO: MSCOCO, CS: Cityscapes, VO: Pascal VOC 2007 CI: Cifar100, CE: CelebA, CB: CUB-200-2011, AG: Agriculture-Vision, US: Kaggle Ultrasound, NI: NIH Chest X-ray. OI: Open Images, OF: Oxford Flowers-101, SV: SVHN.}
    \label{fig:datasets}
    \vspace{-1em}
\end{figure*}

Our experiments span across different image domains including natural images~\cite{OpenImages,deng2009imagenet,pascal-voc-2007,cordts2016cityscapes,lin2014microsoft,krizhevsky2009learning}, X-ray~\cite{wang2017chestx}, ultrasound~\cite{ultrasound}, and satellite images~\cite{chiu2020agriculture}. We also experiment with models in various computer vision tasks including image classification, semantic segmentation and object detection. The datasets and corresponding example images are illustrated in Figure~\ref{fig:datasets}. We train floating-point models on all the datasets from Figure~\ref{fig:train_start} to Figure~\ref{fig:train_end}, where each dataset has one or more associated models. These datasets and three additional datasets from Figure~\ref{fig:calib_start} to Figure~\ref{fig:calib_end} are used as calibration sets. When calibration and training data have different sizes, we resize the calibration data to the same size of the training data. One exception is when calibrating ResNet-18 on Cifar100, calibration samples are randomly cropped. In addition, when Agriculture-Vision data is used to calibrate models on other datasets, only RGB channels are used. When other datasets are used to calibrate Agriculture-Vision models, we use RGB channels to compute a grayscale channel to be used as NIR channel.

These models to be quantized are summarized in Table~\ref{tab:models}. The evaluation metrics are summarized as below:
\begin{enumerate}
    \item All classification tasks except CelebA: top-1 accuracy.
    \item CelebA: average top-1 accuracy on 40 attributes.
    \item Pascal VOC 2007: mAP.
    \item MSCOCO, Cityscapes and Agriculture-Vision: mIoU.
    \item Ultrasound: DICE.
\end{enumerate}
Specifically, DICE is a widely used evaluation metric for binary segmentation, which is defined as
\begin{equation}
    Dice = \frac{2|\bm{p}\cap \bm{g}|}{|\bm{p}| + |\bm{g}|},
\end{equation}
where $\bm{p}$ and $\bm{g}$ are binary prediction and ground truth tensors respectively.

\begin{table}[t]
    \centering
    \footnotesize
    \begin{tabular}{c|c|c}
        \hline
        Datasets & Tasks & Models \\
        \hline
        \multirow{2}{*}{Imagenet} & \multirow{2}{*}{C} & ResNet-18, ResNet-50~\cite{he2016deep} \\ 
         & & Inception-V3~\cite{szegedy2016rethinking} \\
        MSCOCO & S & FCN~\cite{long2015fully} \\
        Cityscapes & S & BiSeNet~\cite{yu2018bisenet} \\
        Pascal VOC 2007 & D & MobileNetV2 SSD-Lite~\cite{sandler2018mobilenetv2}~\cite{liu2016ssd} \\
        Cifar100 & C & ResNet-18 \\
        CelebA & C & ResNet-50 \\
        CUB-200-2011 & C & MMAL-Net~\cite{zhang2020threebranch} \\
        Agriculture-Vision & S & MSCG-Net-101~\cite{Liu_2020_CVPR_Workshops} \\
        Ultrasound & S & U-net \\
        NIH Chest X-ray & C & ResNet-50 \\
        \hline
    \end{tabular}
    \caption{Training datasets and the corresponding pretrained models. C: Image Classification. S: Semantic Segmentation. D: Object Detection.}
    \label{tab:models}
    \vspace{-1em}
\end{table}

\subsection{Cross-domain Calibration}
\begin{figure*}
    \centering
    \includegraphics[width=0.245\linewidth]{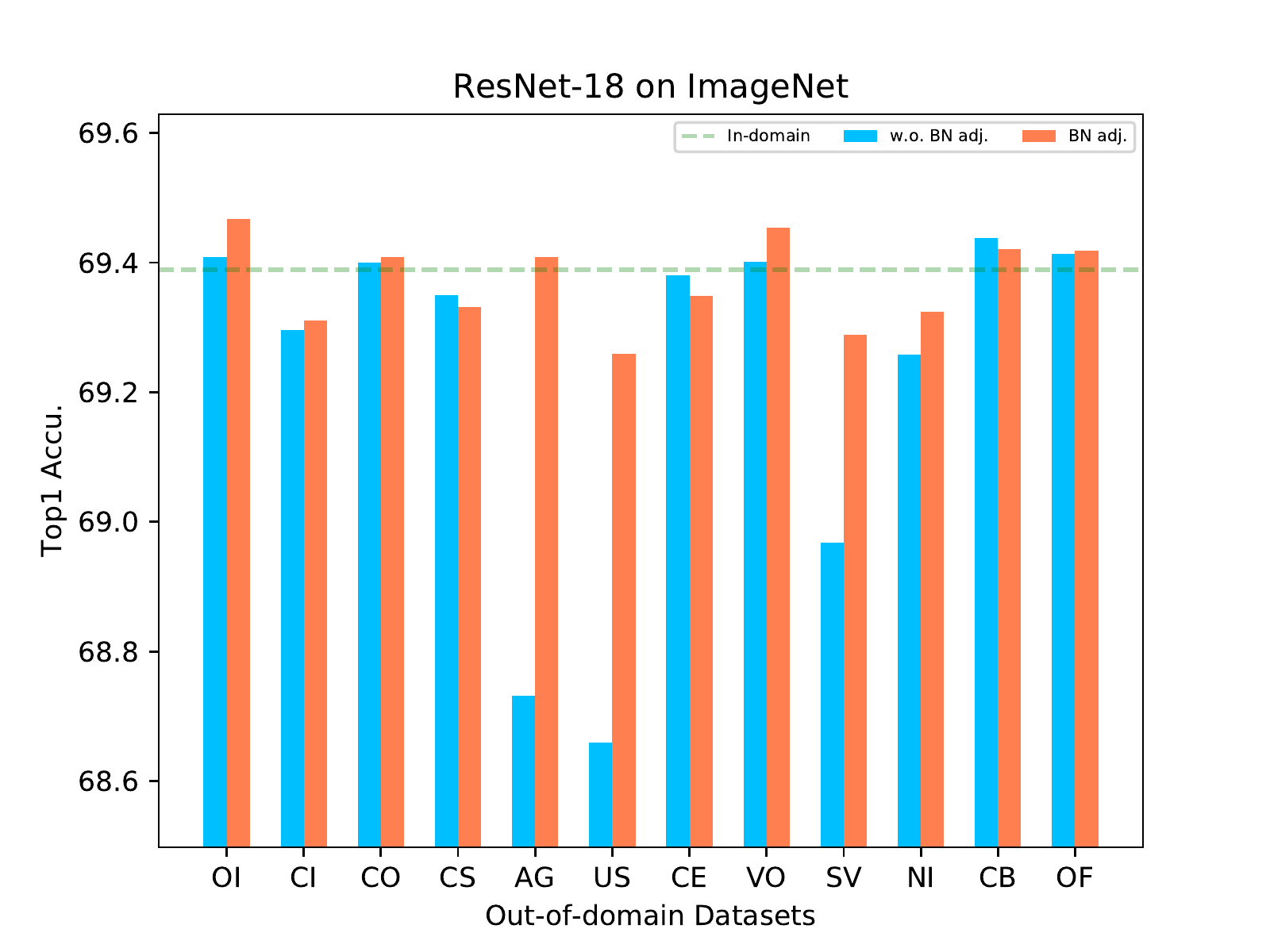}
    \includegraphics[width=0.245\linewidth]{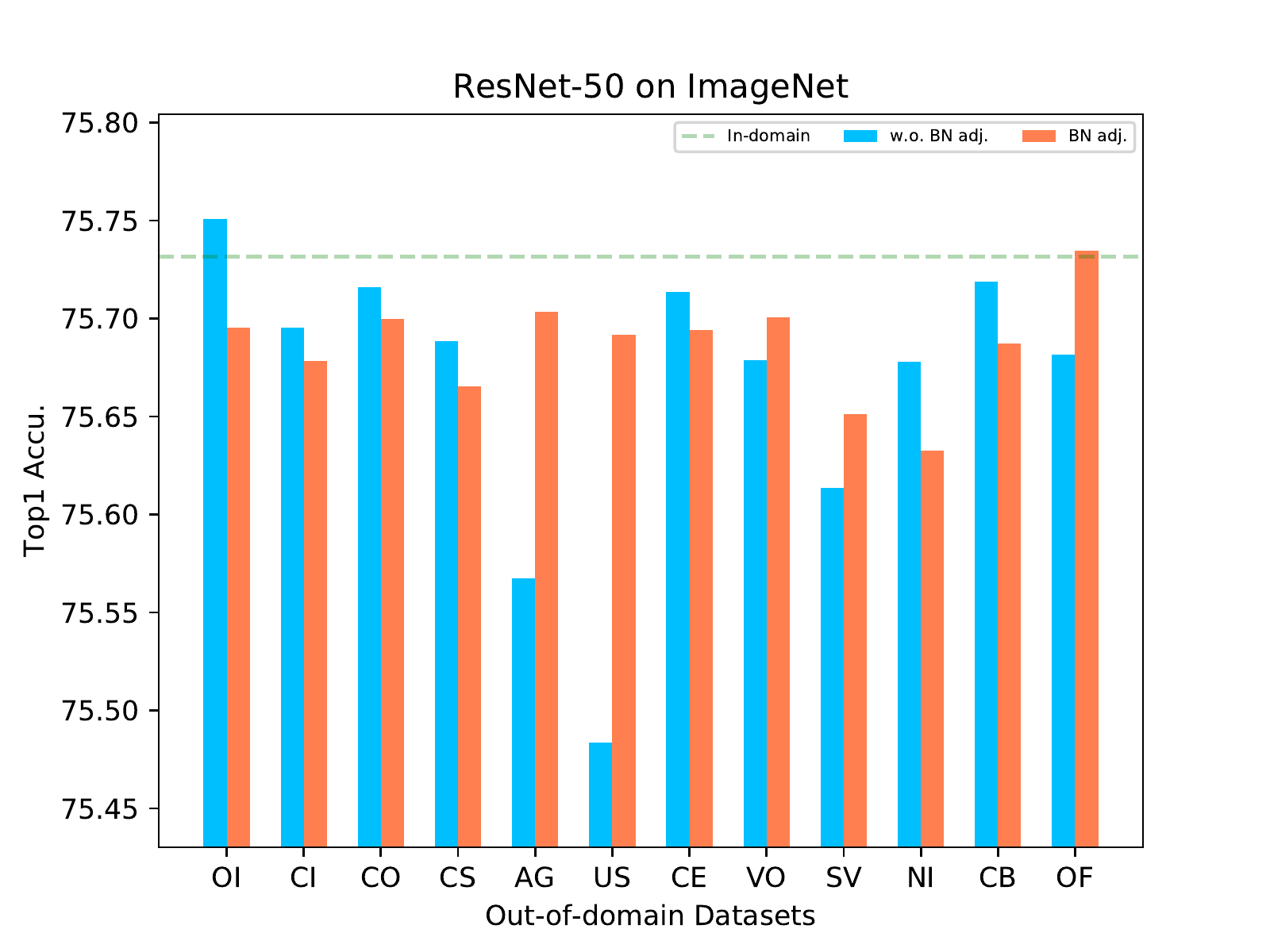}
    \includegraphics[width=0.245\linewidth]{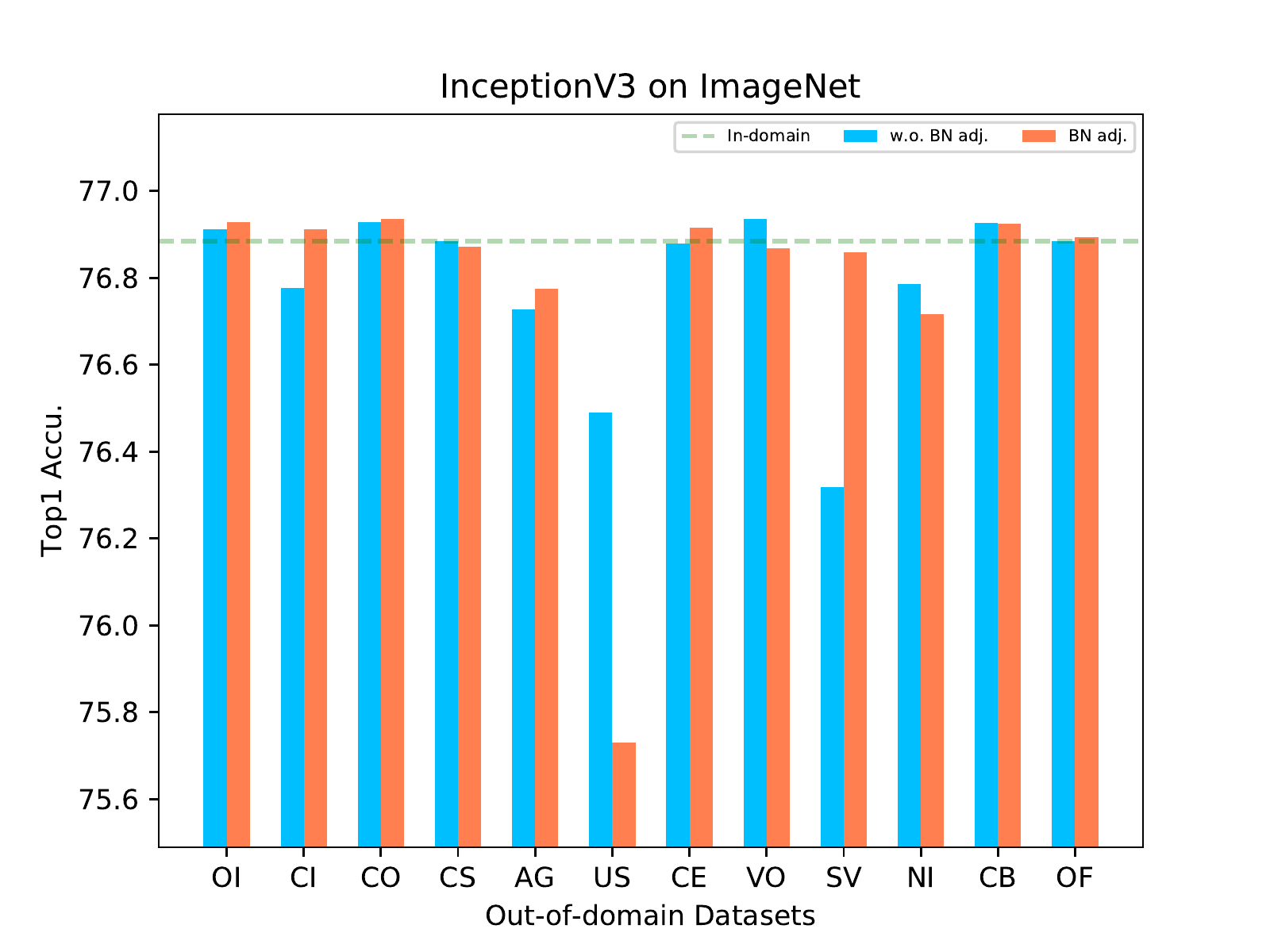}
    \includegraphics[width=0.245\linewidth]{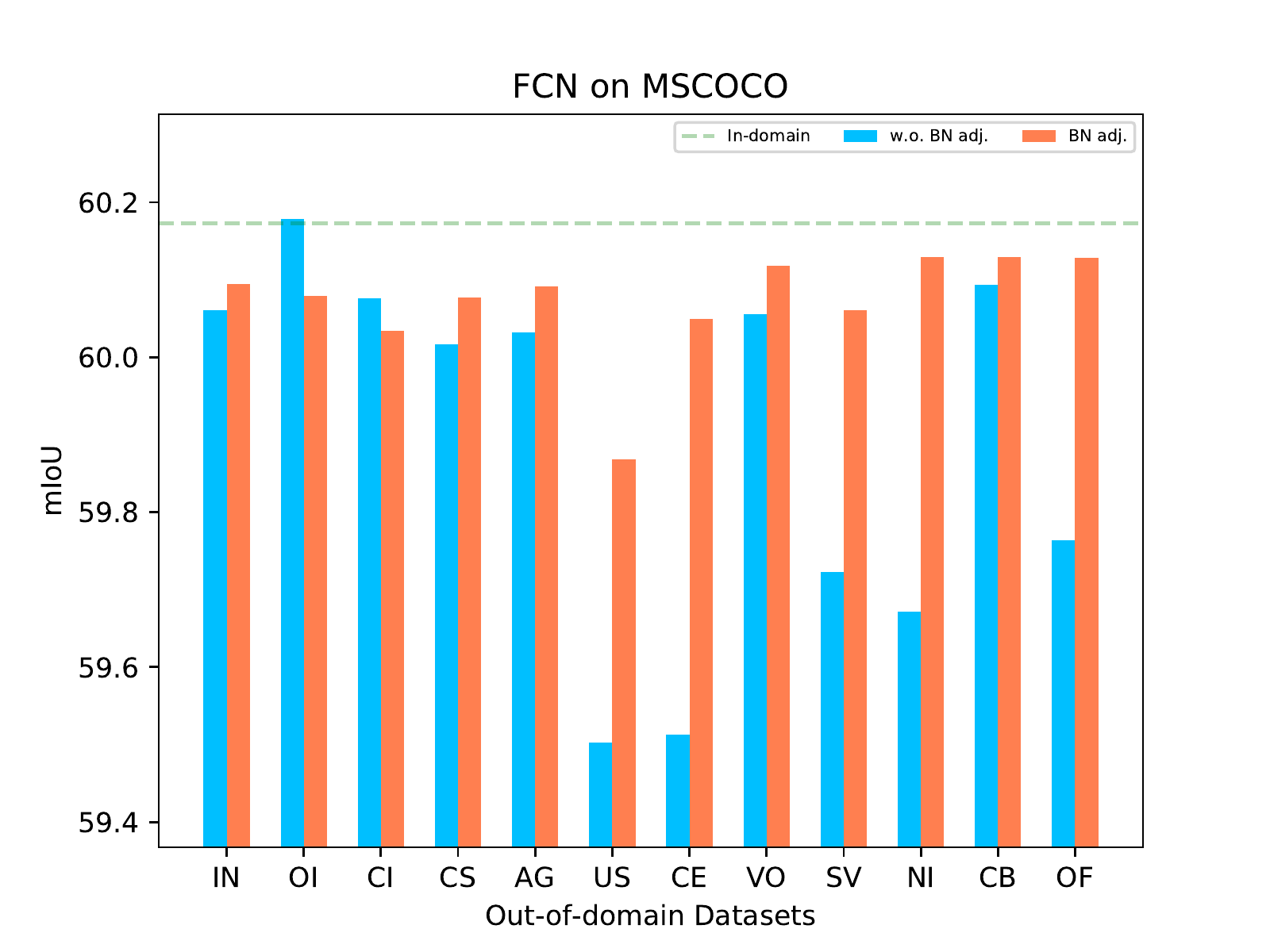}
    \includegraphics[width=0.245\linewidth]{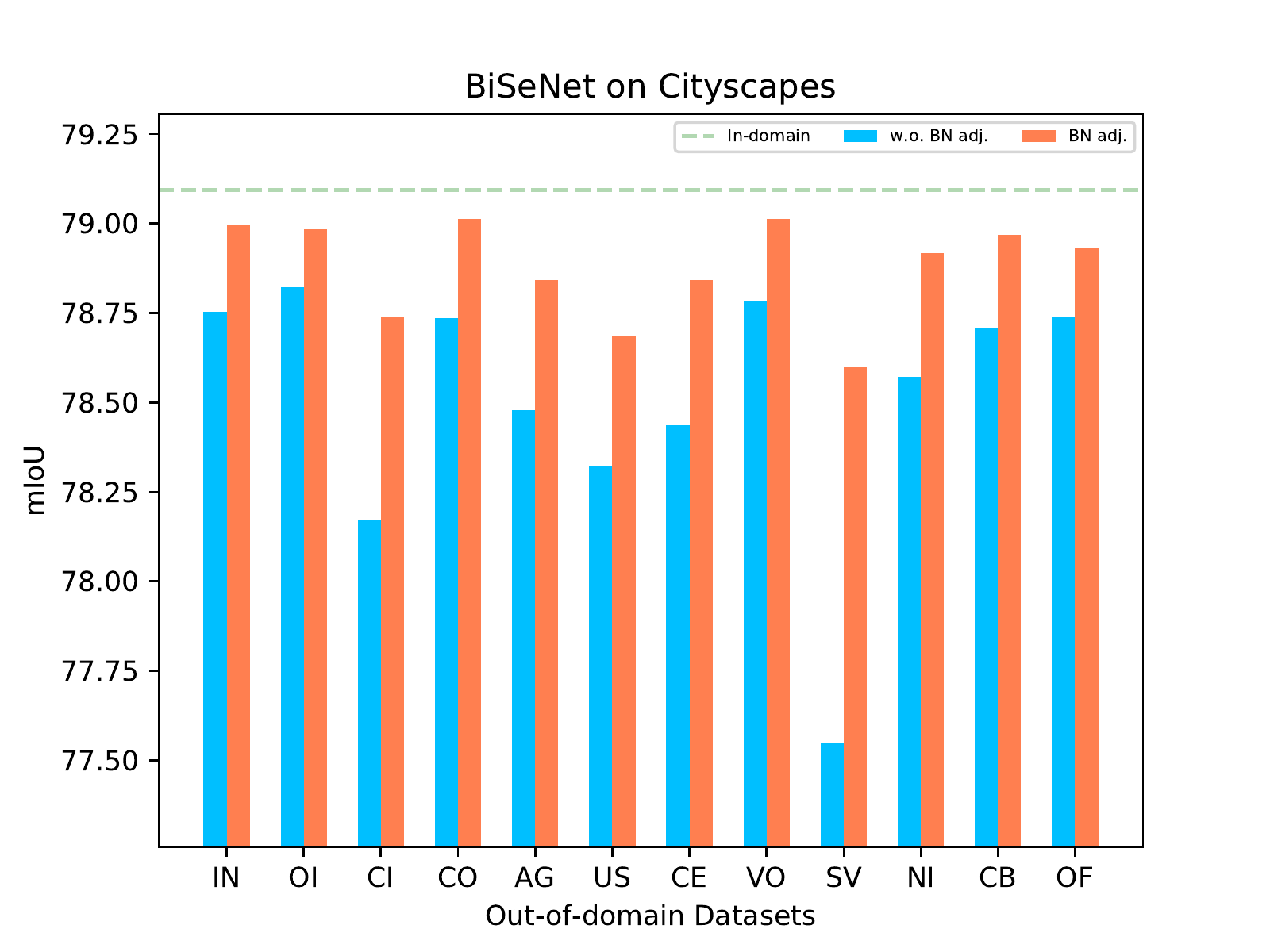}
    \includegraphics[width=0.245\linewidth]{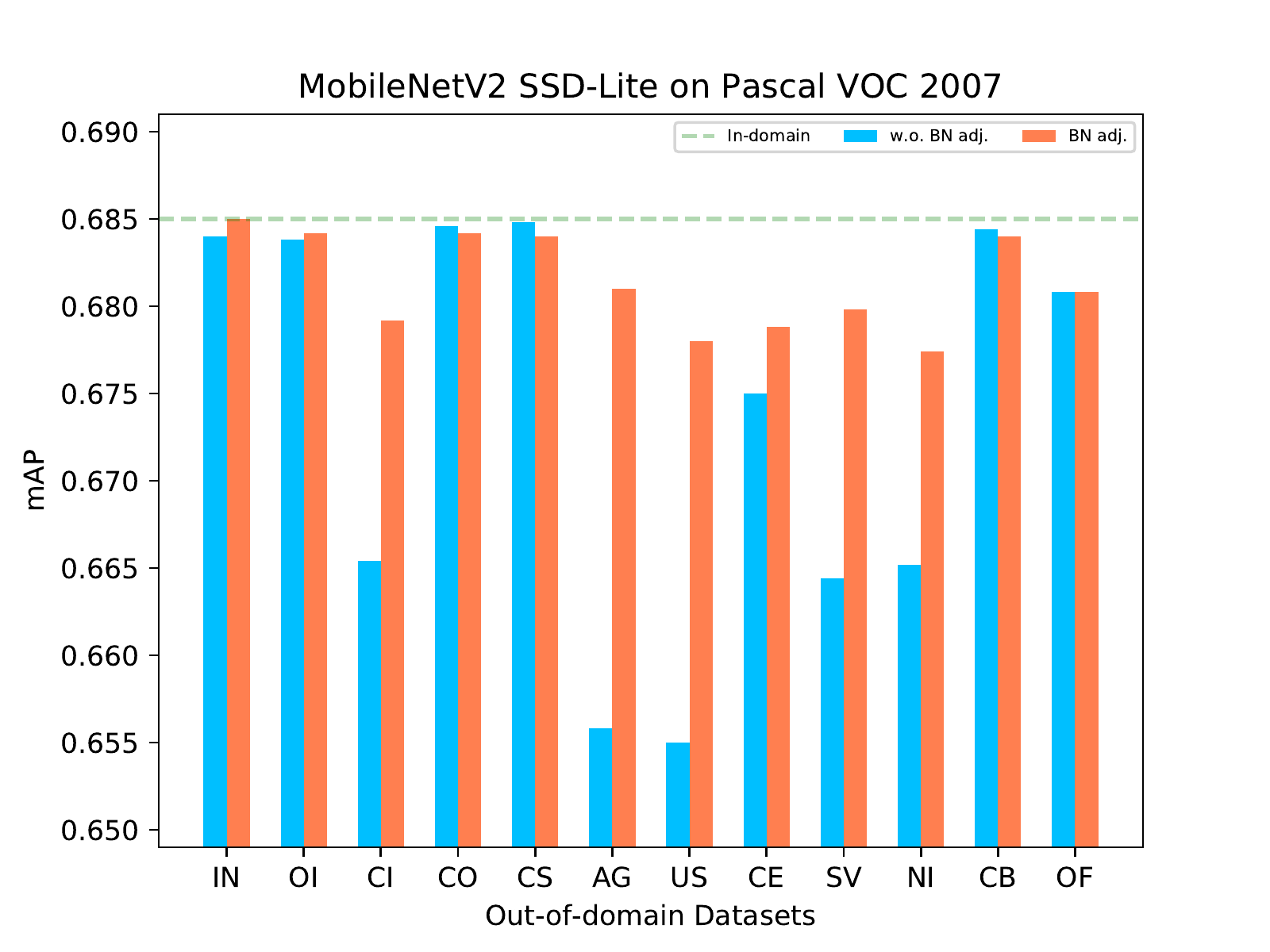}
    \includegraphics[width=0.245\linewidth]{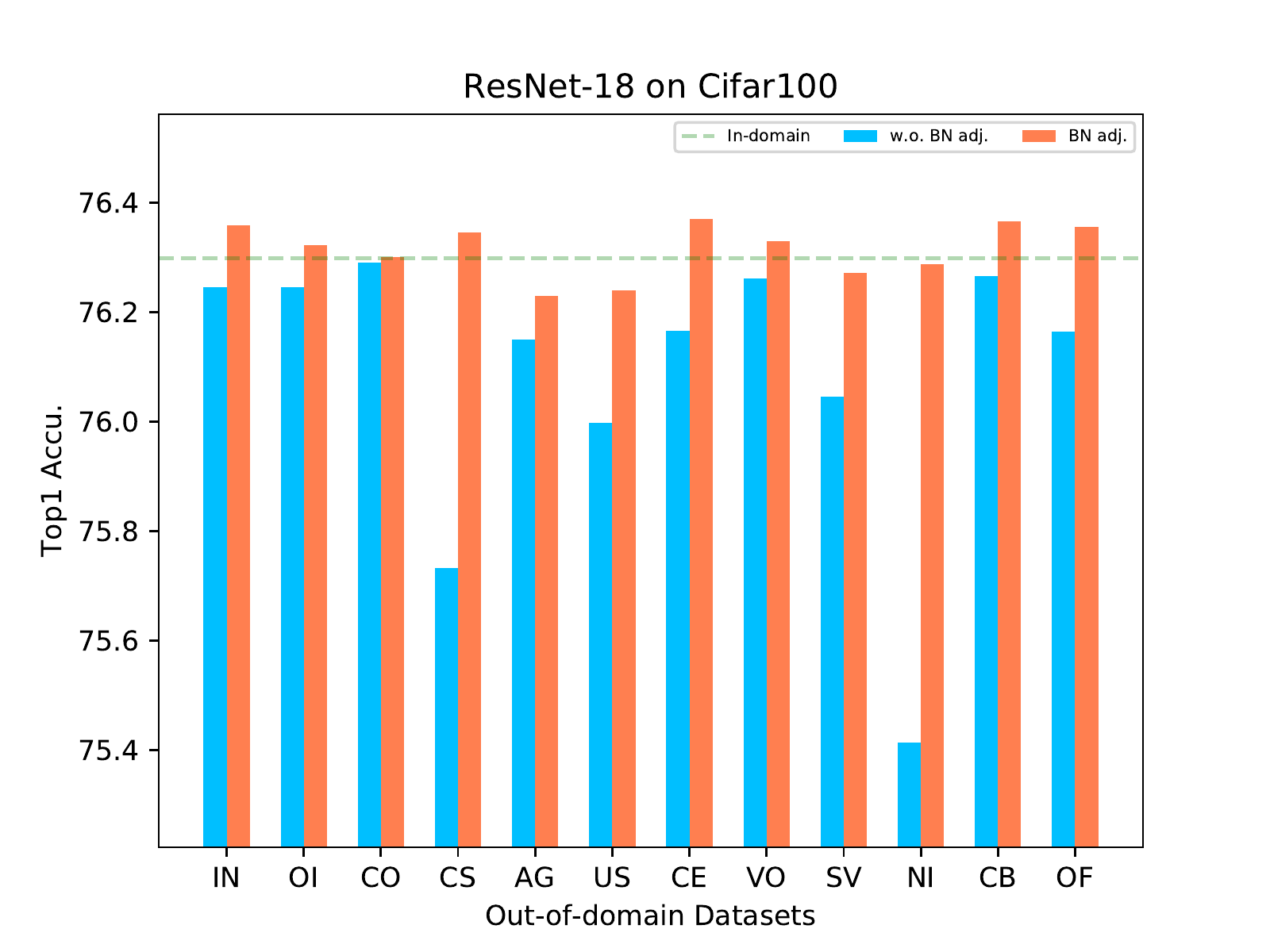}
    \includegraphics[width=0.245\linewidth]{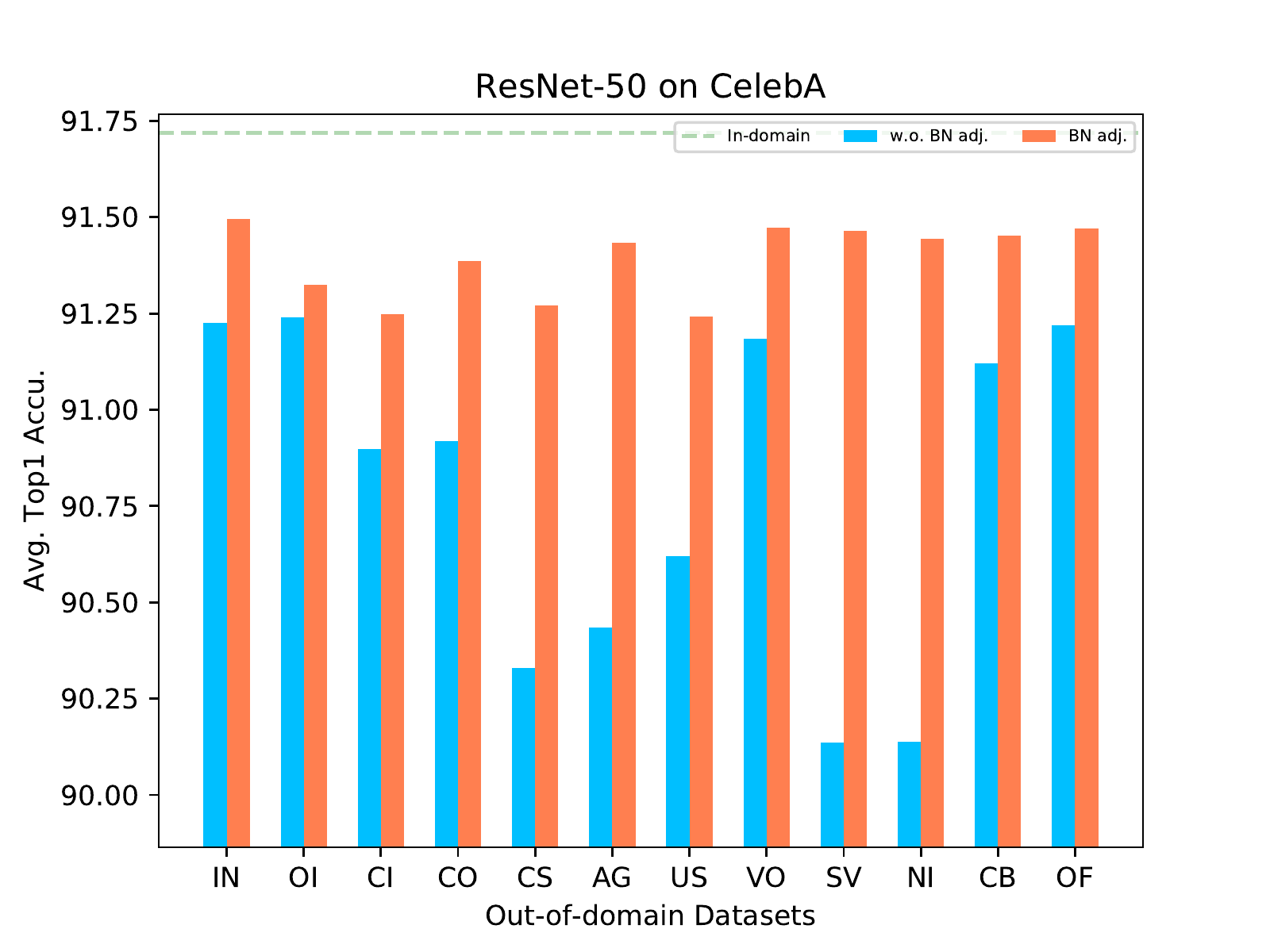}
    \includegraphics[width=0.245\linewidth]{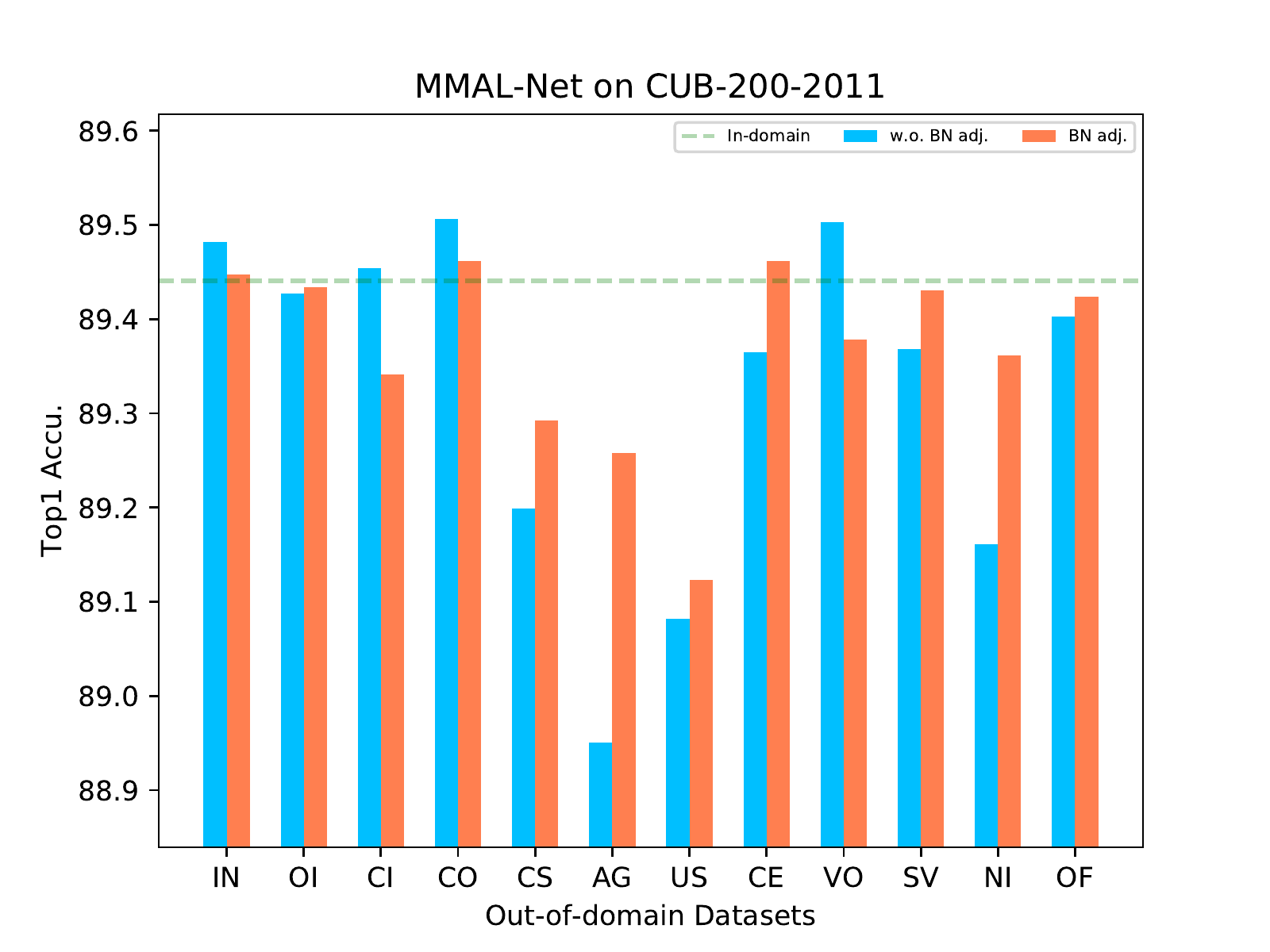}
    \includegraphics[width=0.245\linewidth]{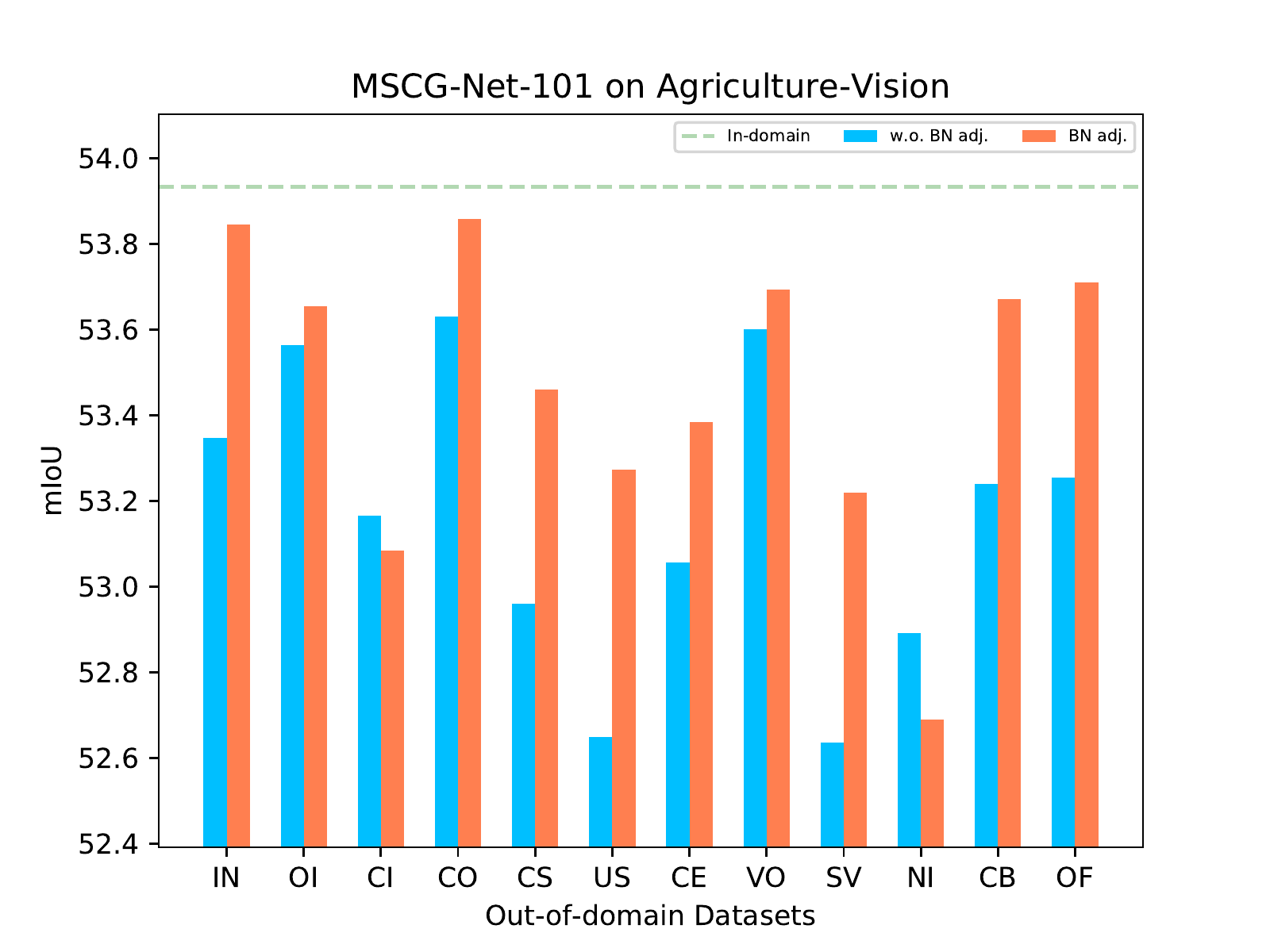}
    \includegraphics[width=0.245\linewidth]{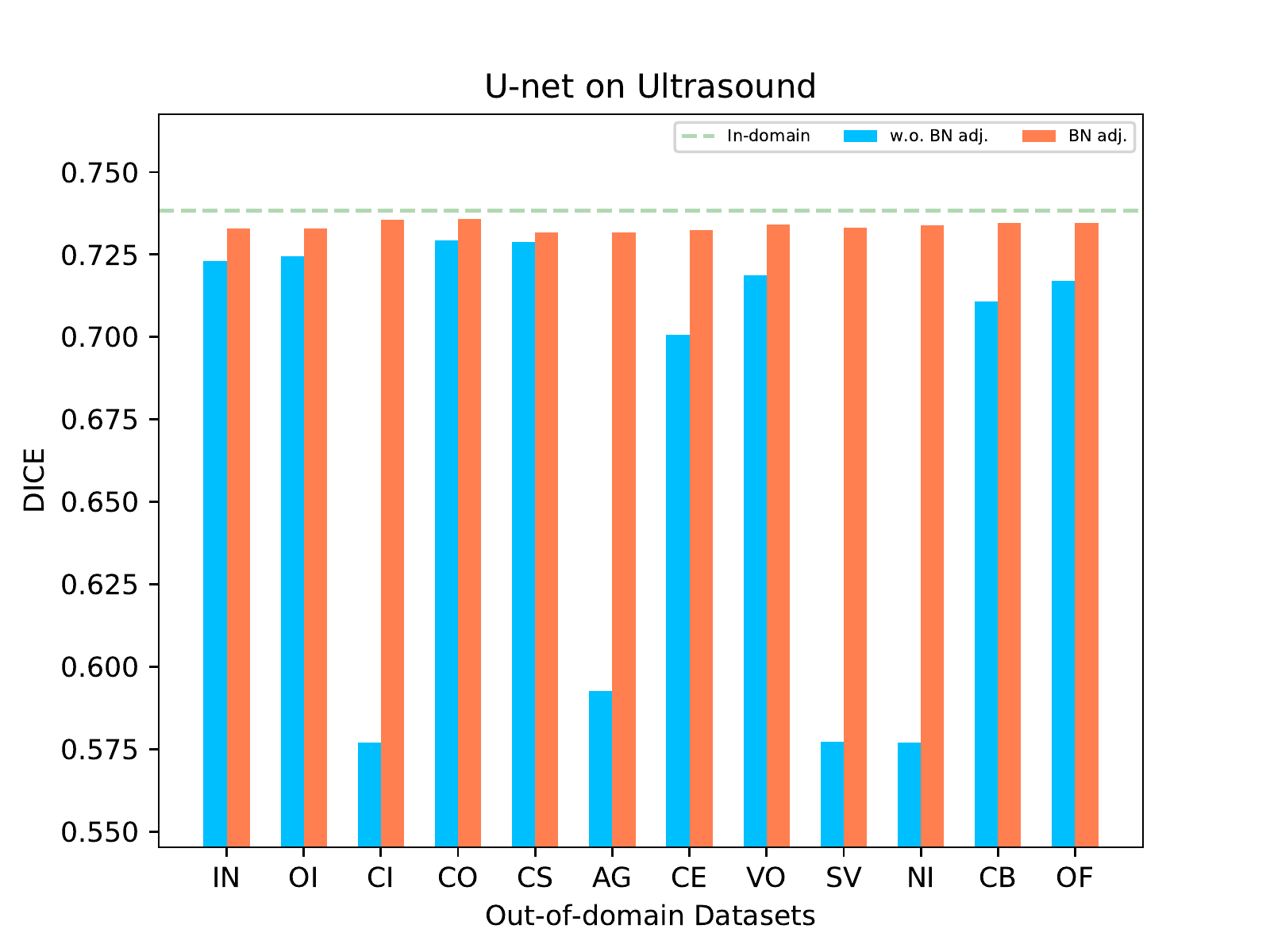}
    \includegraphics[width=0.245\linewidth]{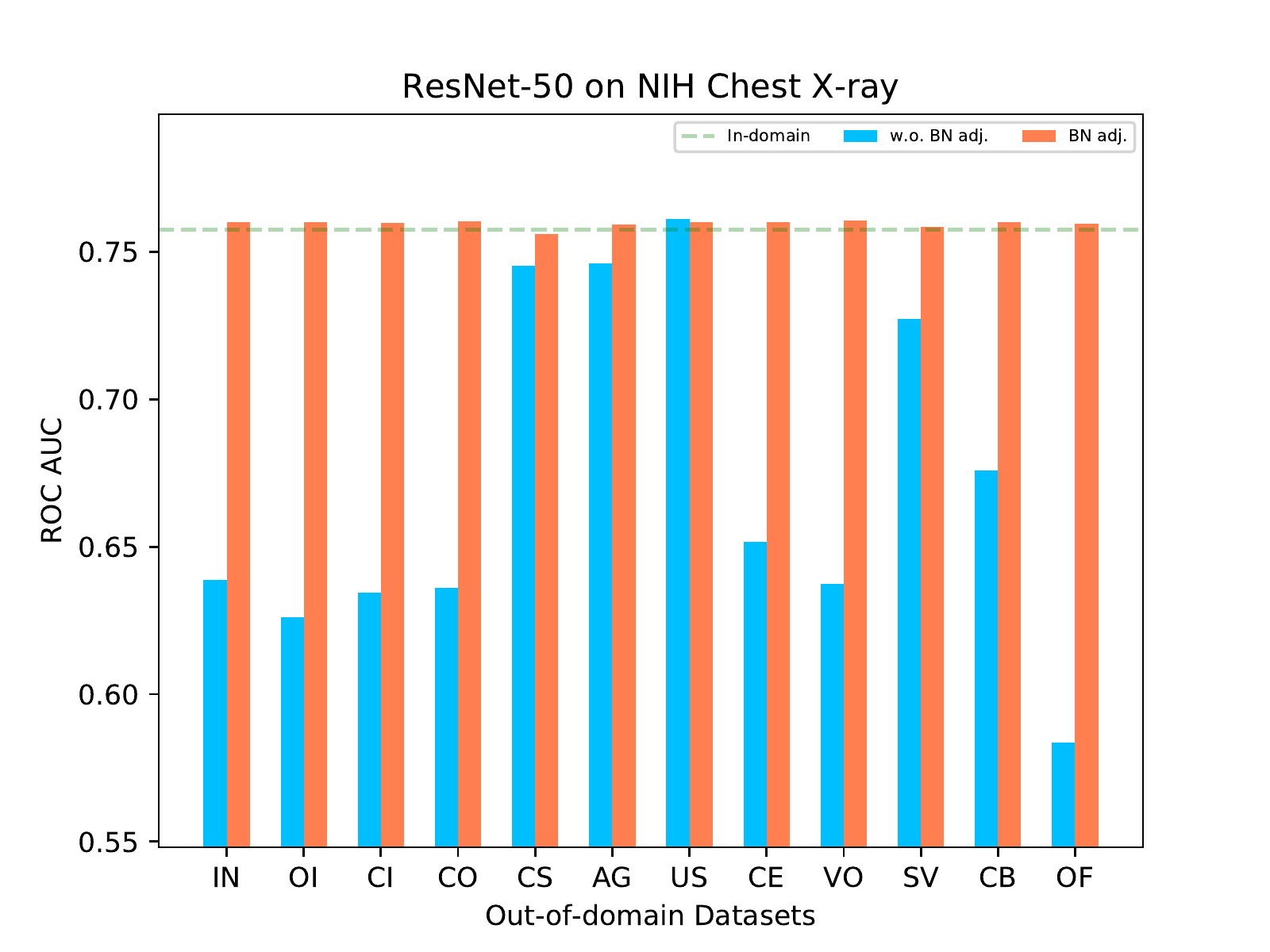}
    \caption{Cross-domain calibration with and without BatchNorm adjustment. Dashed {\color{green}green} line denotes performance of in-domain calibration. The dataset abbreviations are defined in Figure~\ref{fig:datasets}. Zoom in for details.}
    \label{fig:bn_or_not}
    \vspace{-1em}
\end{figure*}

\paragraph{A Naive Approach}
First we employ a naive cross-domain calibration approach, \ie, directly using out-of-domain data to calibrate a model trained in another domain. The results of 8-bit quantization are plotted as {\color{SkyBlue}blue} bars in Figure~\ref{fig:bn_or_not}. The in-domain calibration results are marked in dashed lines.

We can see that for all the datasets and models, a large number of out-of-domain calibration results are comparable to the in-domain baselines. Take ResNet-50 on ImageNet as an example, the gap of out-of-domain and in-domain calibration results is very small. This observation is surprising since most existing PQ methods assume in-domain calibration data is required to estimate activation ranges of intermediate layers.
For some models such as MobileNetV2 SSD-Lite on Pascal VOC, ResNet-50 on CelebA, U-net on Ultrasound, and ResNet-50 on NIH Chest X-ray, there is still an accuracy gap between out-of-domain calibration results and the baseline in-domain results. Such discrepancy often occurs when the original and the calibration domain have a large gap, \eg most natural image datasets are drastically different from ultrasound images. Motivated by this observation, we introduce an approach to reduce the representation gap between the original and the calibration domains for the target model, namely BatchNorm adjustment. 

\paragraph{BatchNorm Adjustment}
Since BatchNorm was proposed to reduce Internal Covariate Shift~\cite{batchnorm-ioffe15}, it has been widely applied to deep neural networks. For one neural network, different domains will generate different BatchNorm parameters. One interesting application is that BatchNorm parameters can be updated on a new domain for domain adaptation~\cite{li2016revisiting,li2017demystifying}. As stated by these work, BatchNorm layers encode information that is specific to the domain that the model is trained on. Inspired by this, we propose BatchNorm adjustment to adapt the models to the out-of-domain calibration datasets. We describe our method as below:
\begin{enumerate}
    \item Given a model $M$ to be quantized and an out-of-domain dataset $D$.
    \item Reset running means and variances of all BatchNorm layers in $M$: $\mu \leftarrow 0$, $\sigma \leftarrow 1$.
    \item Run $M$ on $D$ to accumulate new BatchNorm statistics.
    \item Fold BatchNorm layers to their preceding convolutional or linear layers.
    \item Calculate weight quantization parameters ($s_w$ and $z_w$ in Equation~\ref{eq:sz_weight}).
    \item Run $M$ on $D$ again to calculate activation quantization parameters ($s_a$ and $z_a$ in Equation~\ref{eq:sz_act}).
\end{enumerate}

The results are plotted in {\color{orange}orange} bars in Figure~\ref{fig:bn_or_not}. For most datasets and models, BatchNorm adjustment gets better performance than the baselines without BatchNorm adjustment. In most cases, the results are comparable to or even slightly better than the in-domain calibration results. On ResNet-18/50 on ImageNet, FCN on MSCOCO, ResNet-18 on Cifar100, U-net on Ultrasound and ResNet-50 on NIH Chest X-ray, all out-of-domain calibration results are within 0.3\% performance gap from the in-domain calibration results. On some tasks such as MMAL-Net on CUB-200-2011 and MSCG-Net-101 on Agriculture-Vision, some out-of-domain data still performs relatively lower than others. In such cases, randomly chosen calibration datasets cannot guarantee performance similar to in-domain calibration. In order to fix this issue, we investigate how domain discrepancy affects calibration and how to improve the calibration results.

\subsection{Influence of Domain Discrepancy on Calibration}
\begin{figure*}
    \centering
    \includegraphics[width=0.245\linewidth]{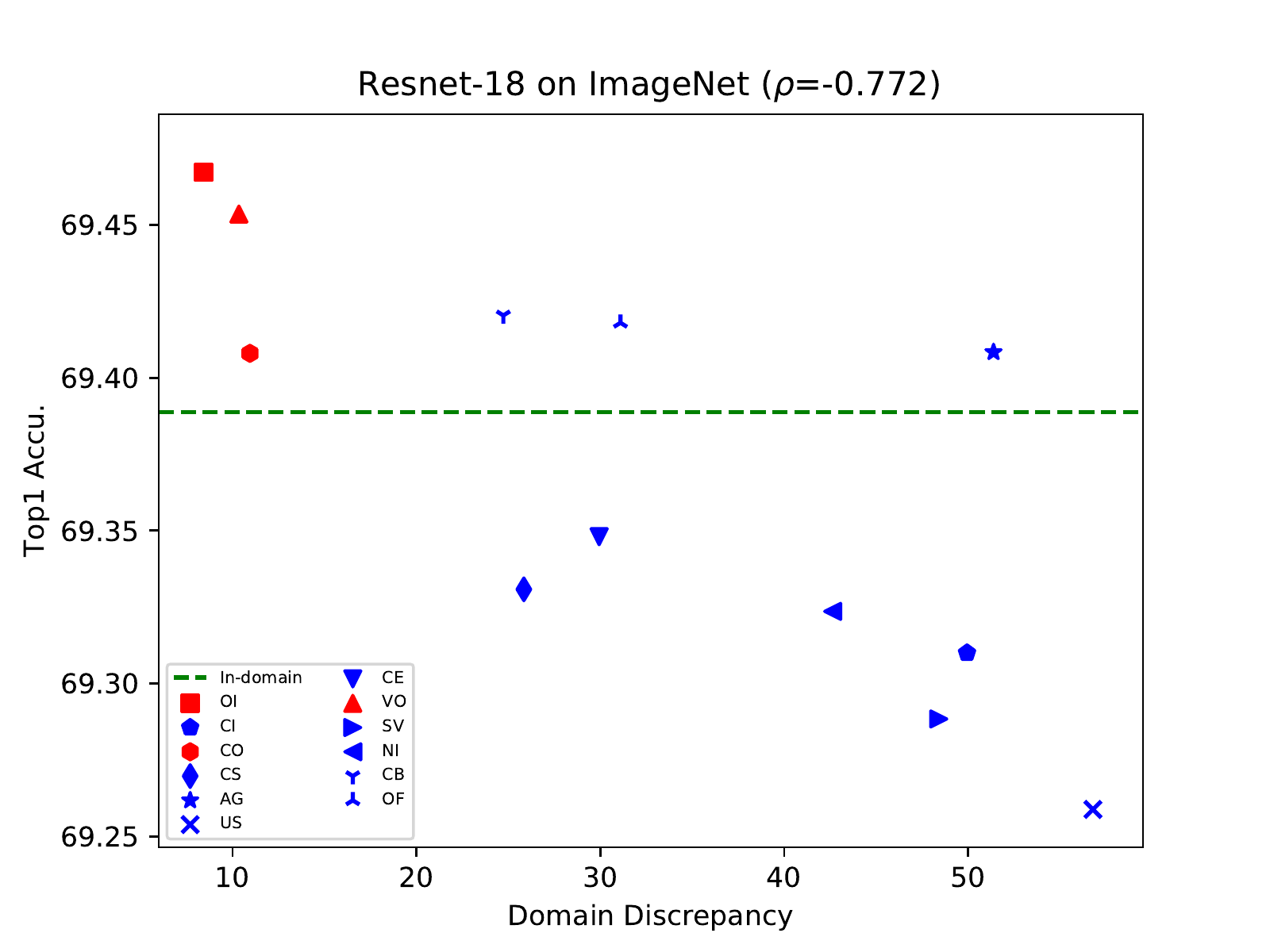}
    \includegraphics[width=0.245\linewidth]{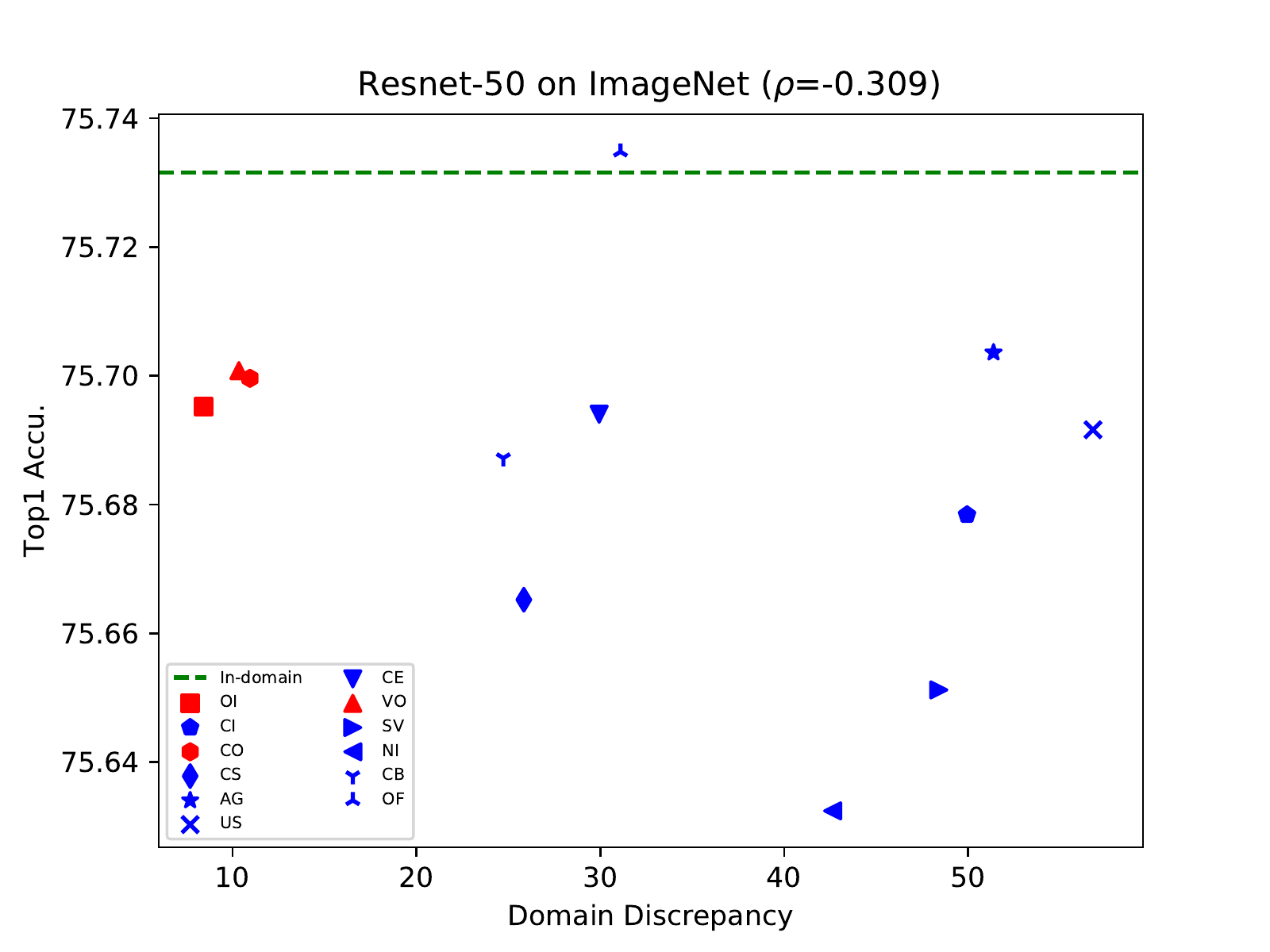}
    \includegraphics[width=0.245\linewidth]{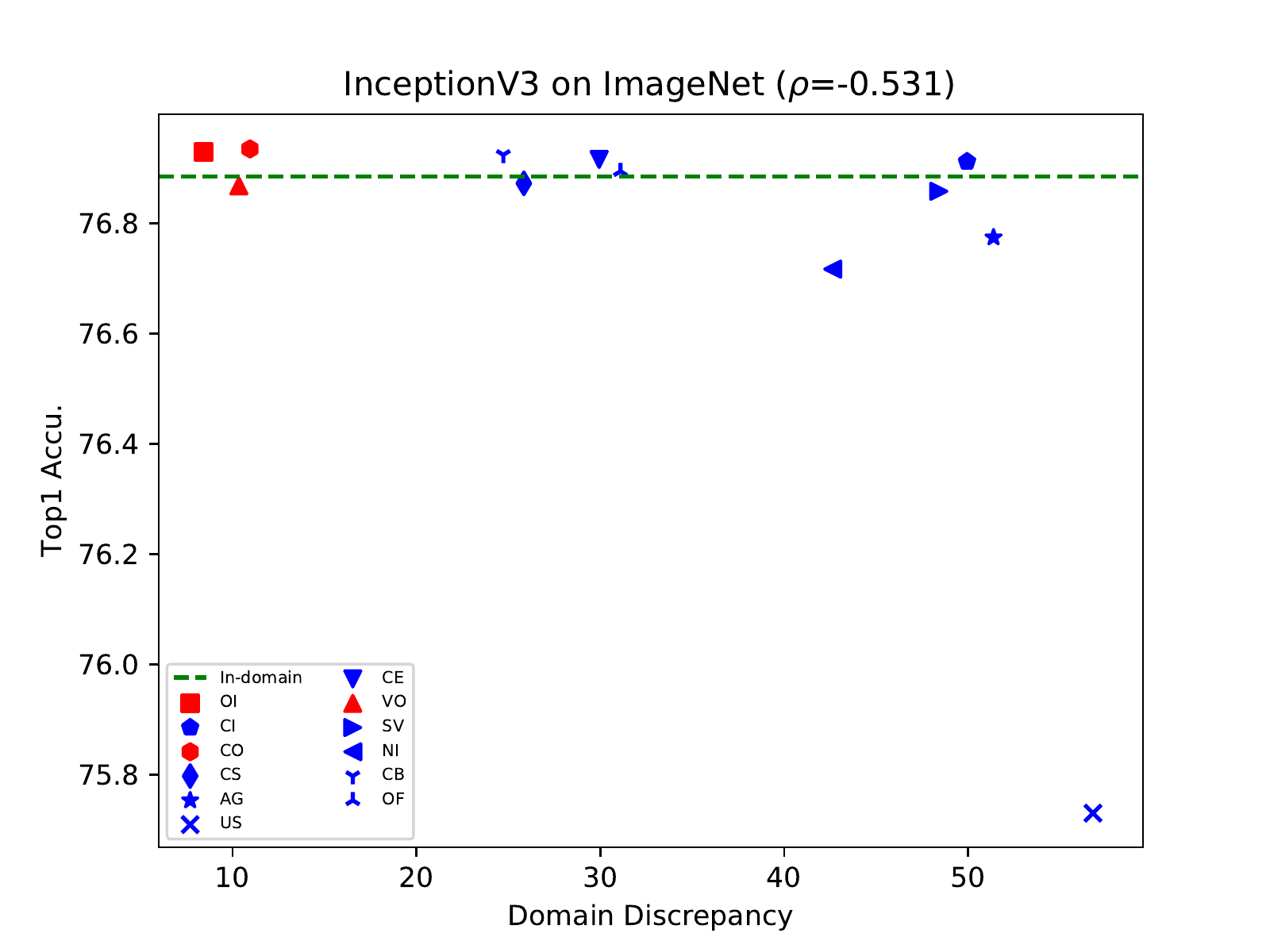}
    \includegraphics[width=0.245\linewidth]{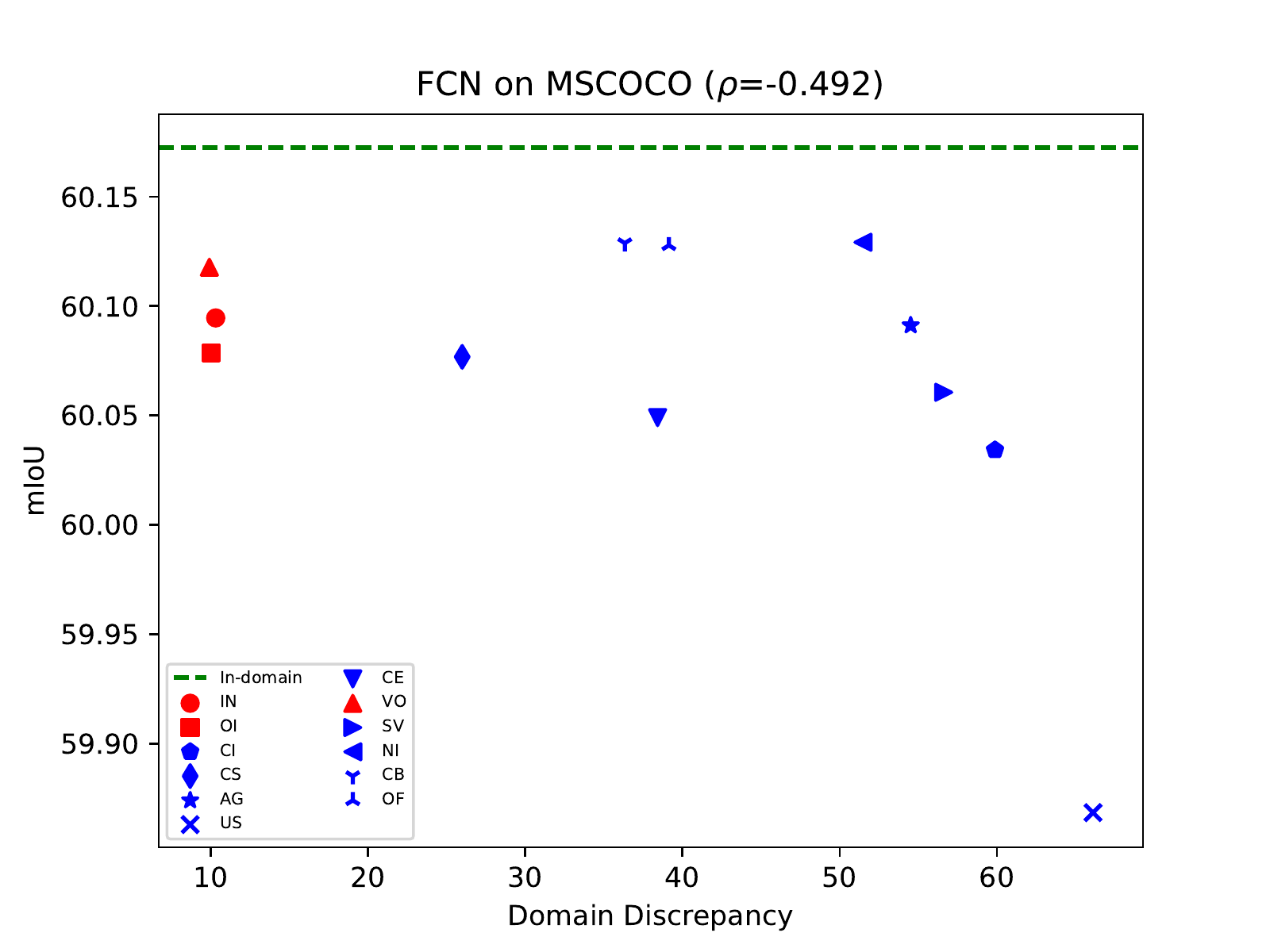}
    \includegraphics[width=0.245\linewidth]{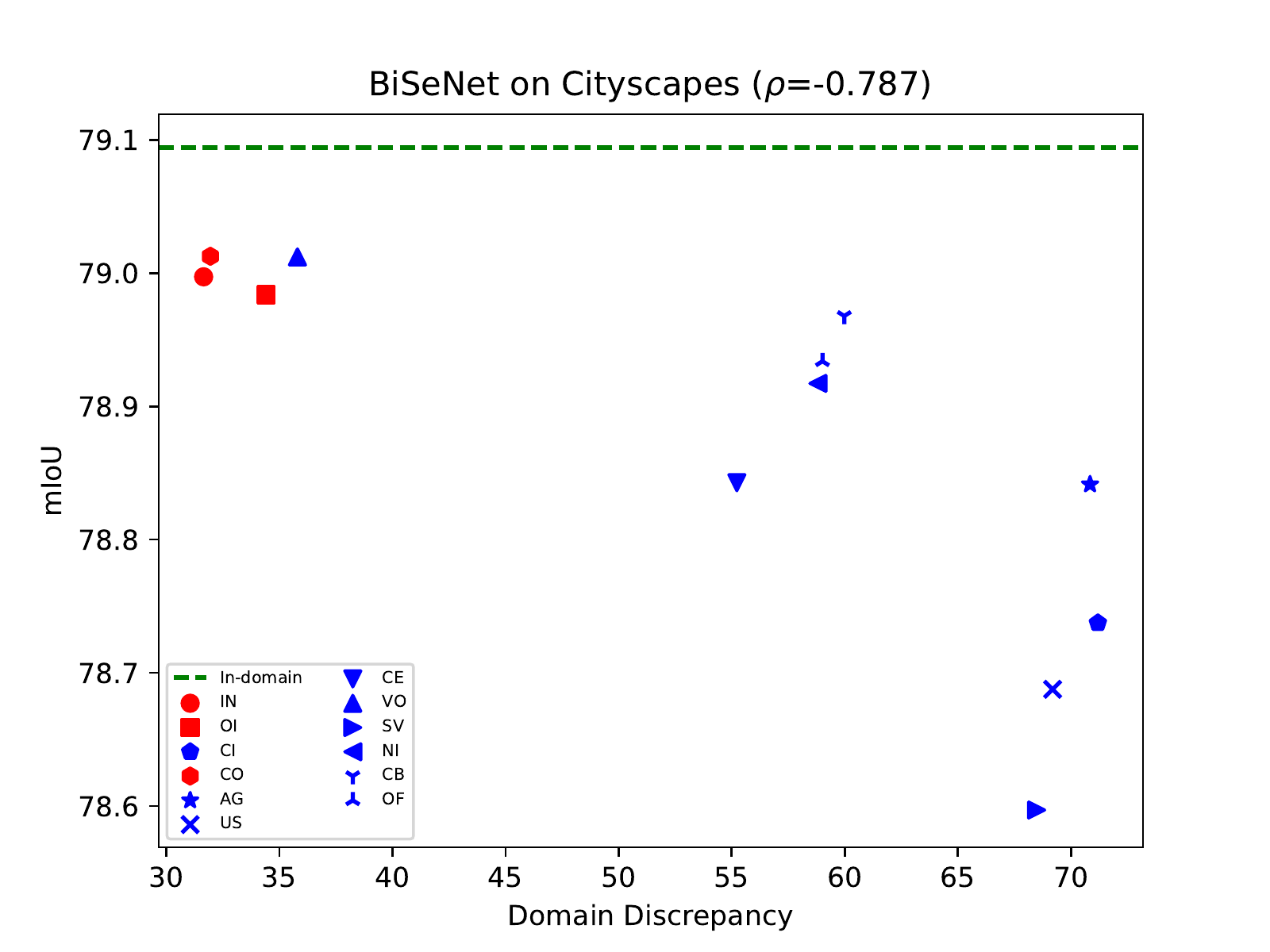}
    \includegraphics[width=0.245\linewidth]{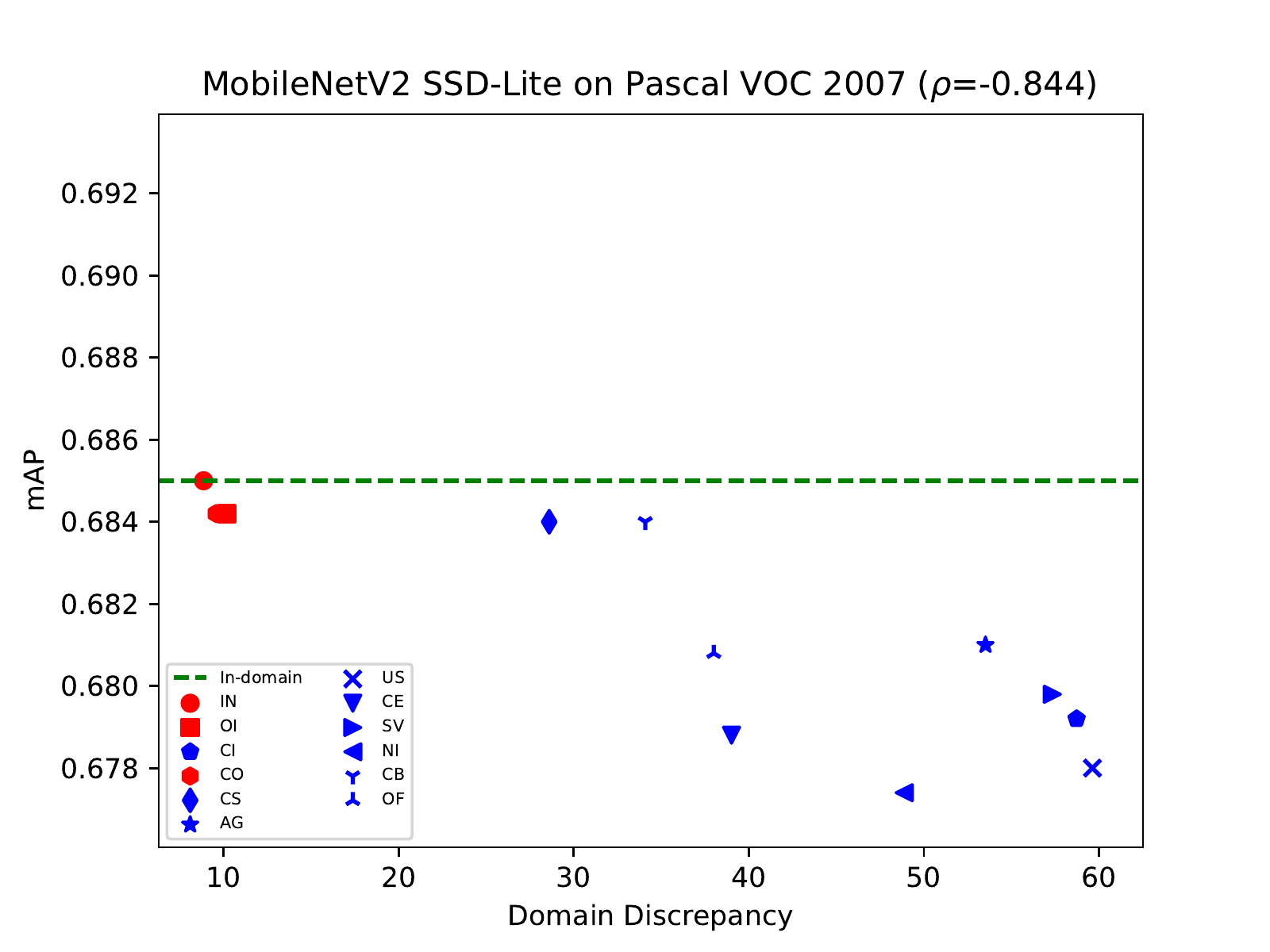}
    \includegraphics[width=0.245\linewidth]{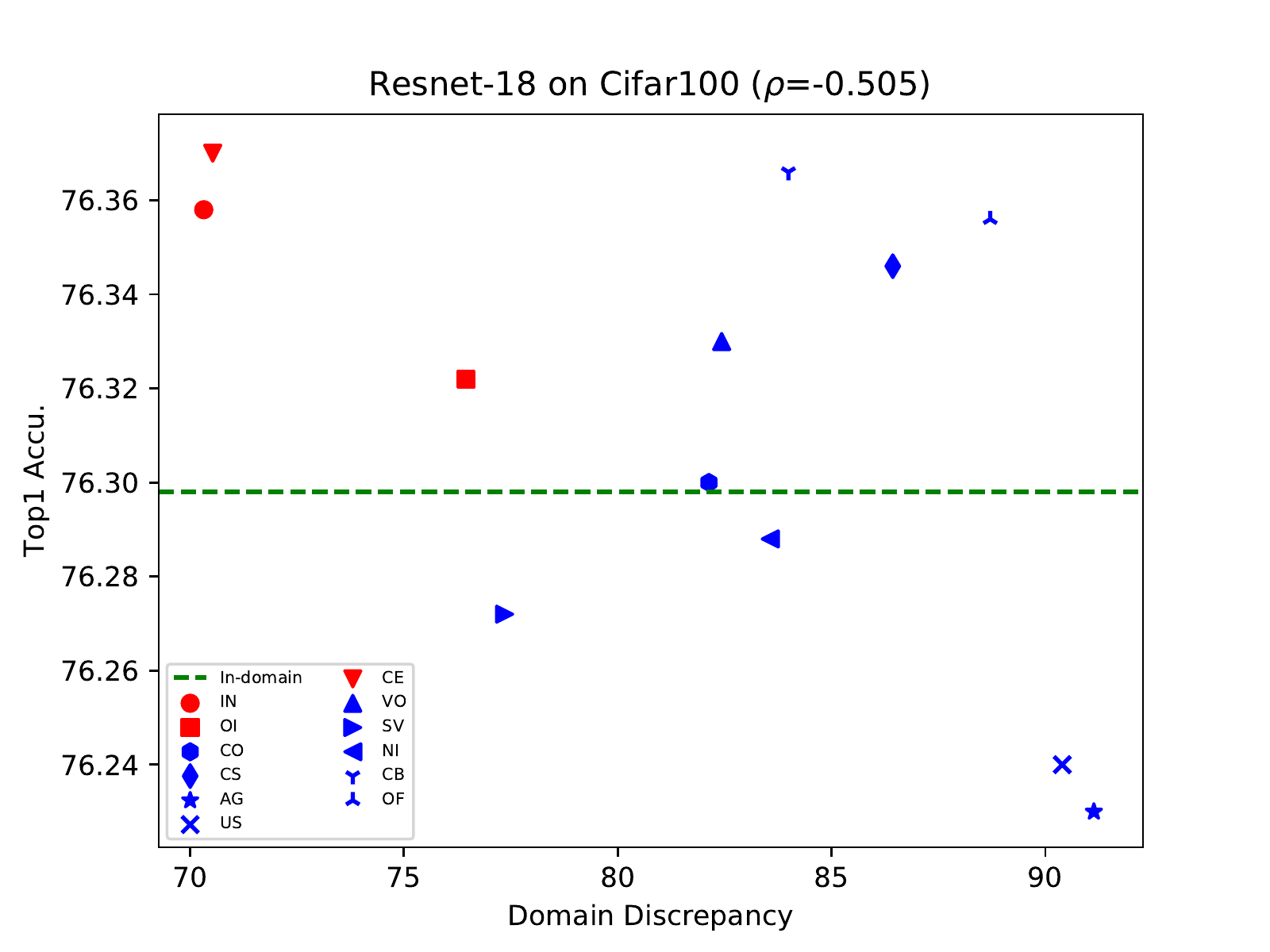}
    \includegraphics[width=0.245\linewidth]{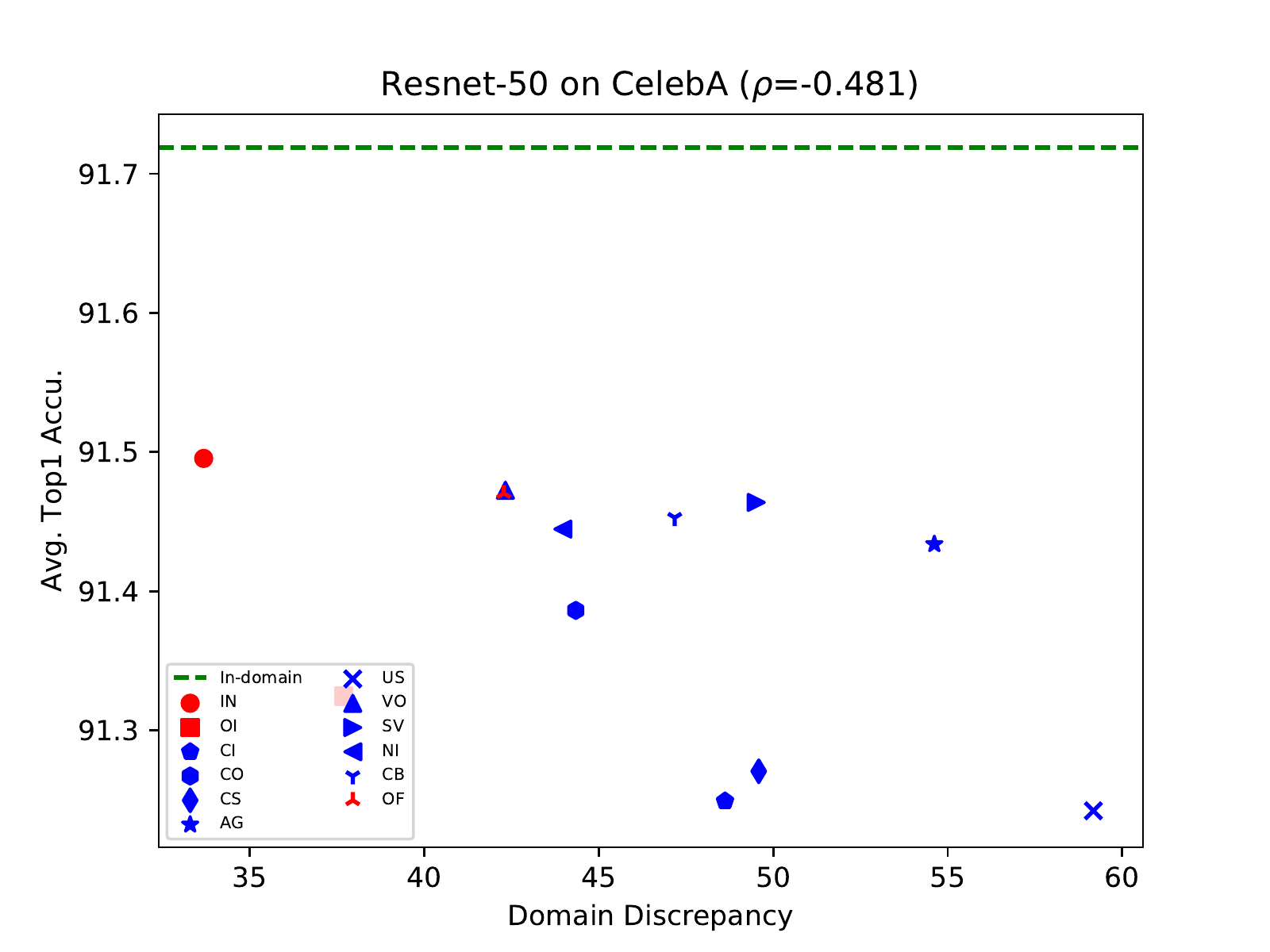}
    \includegraphics[width=0.245\linewidth]{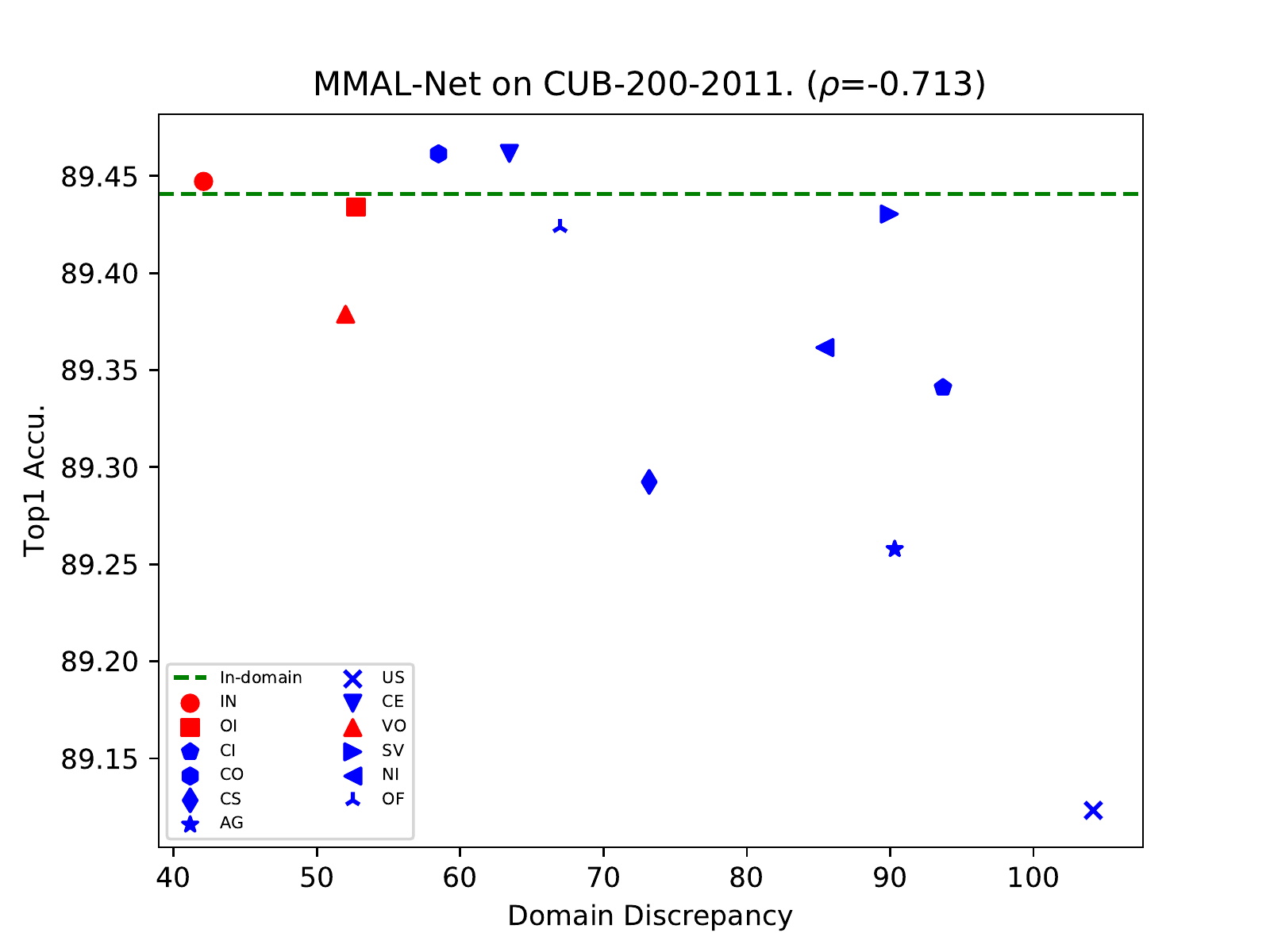}
    \includegraphics[width=0.245\linewidth]{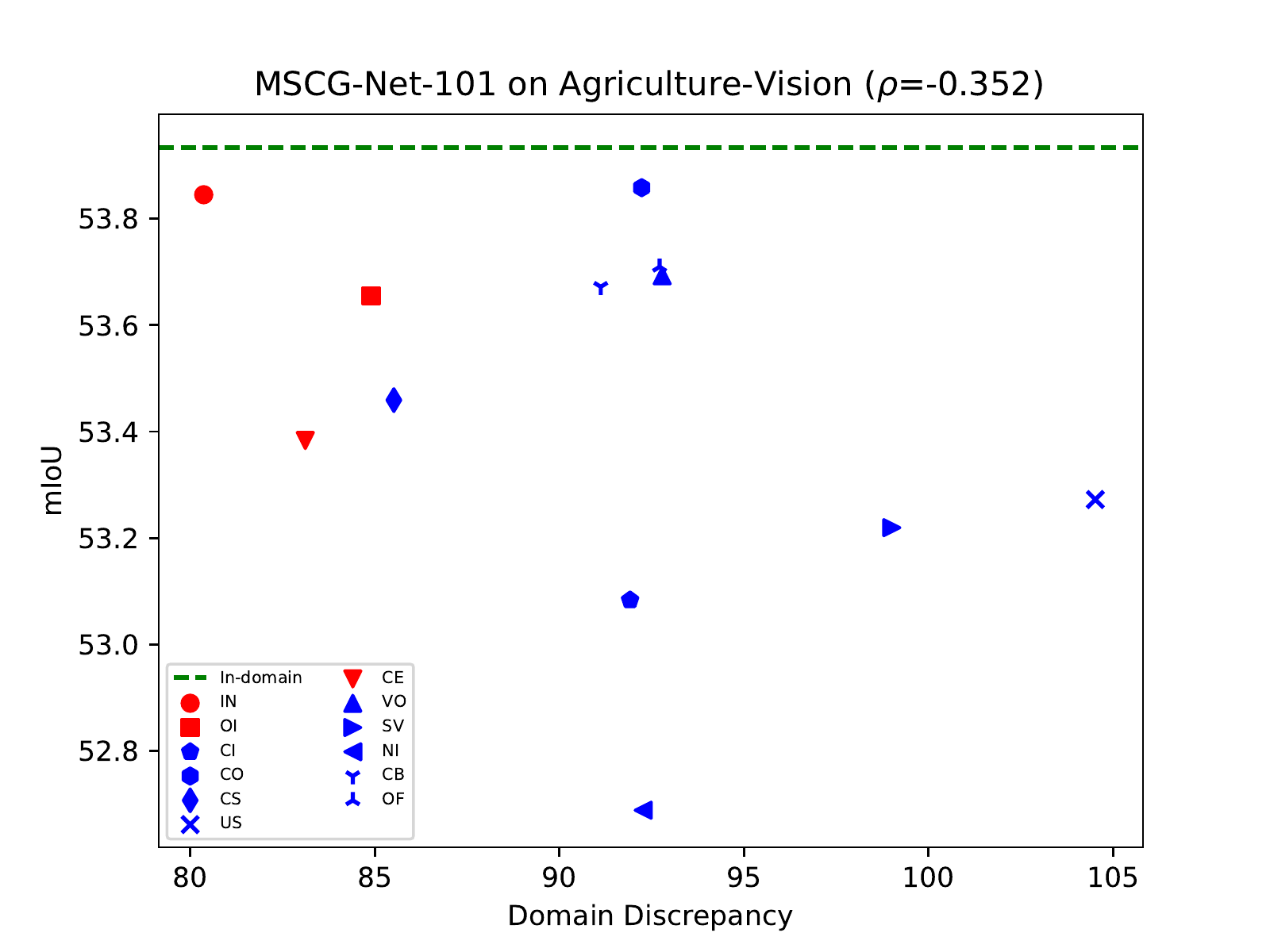}
    \includegraphics[width=0.245\linewidth]{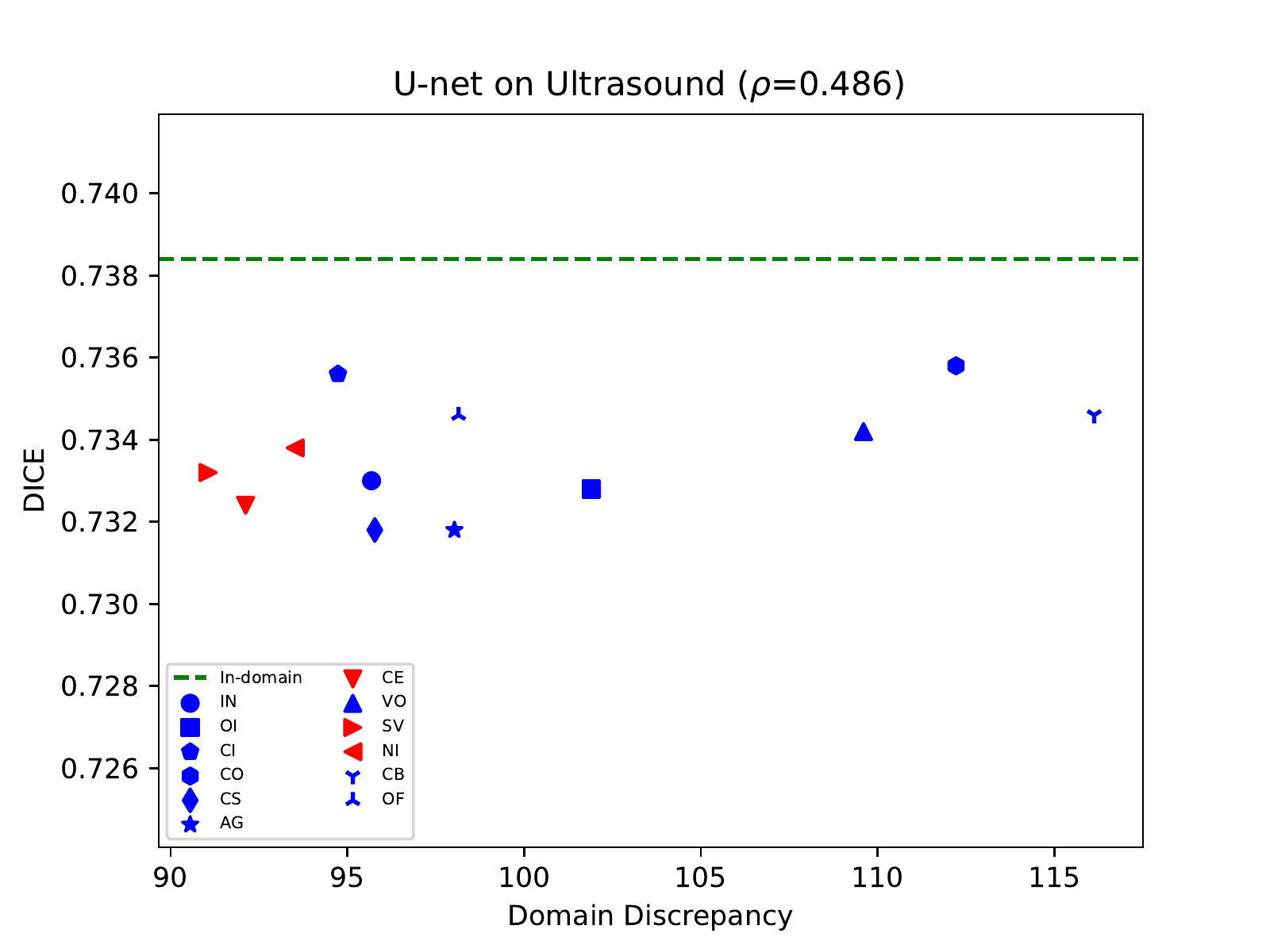}
    \includegraphics[width=0.245\linewidth]{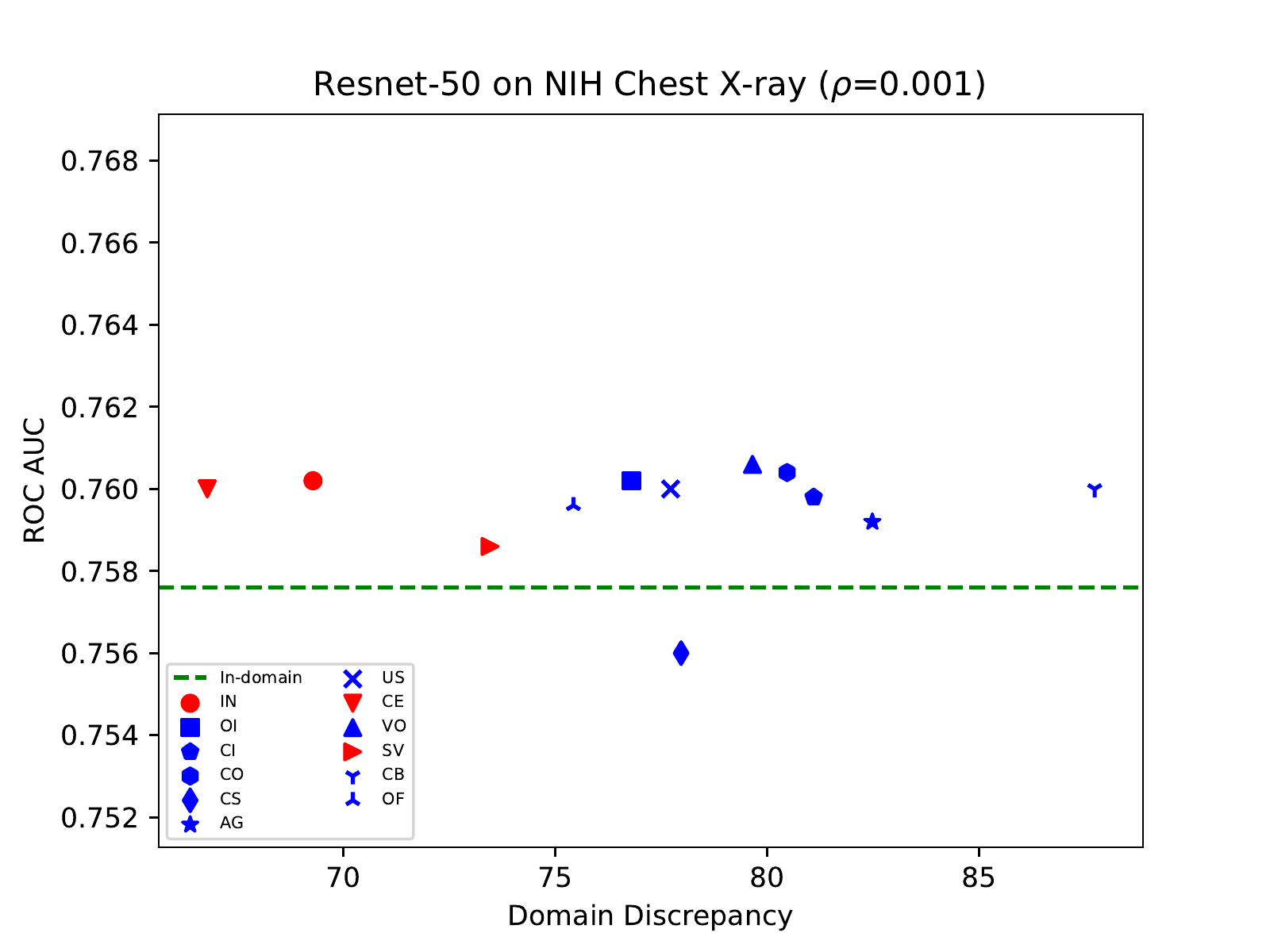}
    \caption{Visualization of correlation between calibration performance and domain discrepancy. Top-3 out-of-domain datasets with smallest domain discrepancy from the in-domain data are marked in {\color{red} red}. Dashed green line denotes performance of in-domain calibration. The dataset abbreviations are defined in Figure~\ref{fig:datasets}. Zoom in for details.}
    \label{fig:score_gram}
    \vspace{-1em}
\end{figure*}

In this section, we investigate the relationship between domain discrepancy and cross-domain calibration performance. Since Maximum Mean Discrepancy (MMD) was proposed to measure difference of sample mean in Reproducing Kernel Hilbert Space~\cite{gretton2012kernel}, it has been widely used as a domain discrepancy measure. As proved by Li~\etal~\cite{li2017demystifying}, matching MMD with the second order polynomial kernel is equivalent to matching Gram matrices of feature maps. Similar to~\cite{li2017demystifying}, we employ mean $L_2$ distance between Gram matrices of feature maps to measure the discrepancy between two image domains. Formally, domain discrepancy $D$ is defined as
\begin{equation}
D = \frac{1}{N^2}\sum_{i=1}^{N}\sum_{j=1}^{N} \lVert \bm{G}_{ij}^{A} - \bm{G}_{ij}^{B}\lVert_{2}^{2},
\end{equation}
where $\bm{G}^{A} \in \mathbb{R}^{N\times N}$ and $\bm{G}^{B} \in \mathbb{R}^{N\times N}$ are average Gram matrices of two domains $A$ and $B$, which are defined as
\begin{equation}
\bm{G}^A = \frac{1}{|A|}\sum_{k=1}^{|A|}\bm{G}_{k}^{A} \quad \textrm{and} \quad \bm{G}^B = \frac{1}{|B|}\sum_{k=1}^{|B|}\bm{G}_{k}^{B},
\end{equation}
where $|A|$ and $|B|$ are number of samples in domain $A$ and $B$ respectively. Ignoring the domain superscript for simplicity, each element $\bm{G}_{k,ij}$ in $\bm{G}_{k}^{A}$ or $\bm{G}_{k}^{B}$ is
\begin{equation}
    \bm{G}_{k,ij} = \sum_{m=1}^{M} \bm{F}_{k,im} \bm{F}_{k,jm},
\end{equation}
where $\bm{F}_{k}$ is the feature embedding of $k$-th sample, which is normalized by mean and standard deviation over all images in this domain, and $M$ is the embedding dimension. In our experiments, we use the output of last conv layer in the fourth conv block \texttt{conv\_4\_3} of VGG-16 to extract the feature maps.

Next we study how domain discrepancy is related to calibration performance. In Figure~\ref{fig:score_gram}, we plot model performance vs domain discrepancy with the source training data into scatter points. We also calculate correlation coefficient between model performance and domain discrepancy shown in the title of each sub-figure. First, on most of the datasets, calibration performance is negatively correlated to gram matrix distance. The out-of-domain datasets that have the smaller domain discrepancy with the in-domain data always achieve comparable performance with in-domain data. We marked the top-3 out-of-domain data with smallest domain discrepancy in red for better visualization.
This observation can be useful for cross-domain calibration, when calibration/training data is sensitive and inaccessible but some high-level statistics such as Gram matrix are available. One can ask the data owner to provide a mean Gram matrix of the source dataset, which can be used to search in the pre-built candidate pool of cross-domain datasets for the out-of-domain calibration dataset with smallest domain discrepancy. 

\begin{table*}[t]
    \small
    \centering
    \begin{tabular}{c|C{0.67cm}|C{0.67cm}|C{0.67cm}|C{0.67cm}|C{0.67cm}|C{0.67cm}|C{0.67cm}|C{0.67cm}|C{0.67cm}|C{0.67cm}|C{0.67cm}|C{0.67cm}}
        \hline
        \mr{Calib Methods \textbackslash~Datasets} & \multicolumn{3}{c|}{IN} & \mr{CO} & \mr{CS} & \mr{VO} & \mr{CI} & \mr{CE} & \mr{CB} & \mr{AG} & \mr{US} & \mr{NI} \\
        \cline{2-4}
         & R18 & R50 & IV3 & & & & & & & & &  \\
        \hline
        FP32 & 69.76 & 76.13 & 77.46 & 60.47 & 79.10 & 0.686 & 76.50 & 91.74 & 89.63 & 54.63 & 0.730 & 0.773 \\
        \hline
        In-domain & 69.39 & 75.73 & 76.88 & 60.17 & 79.09 & 0.685 & 76.30 & 91.72 & 89.36 & 53.93 & 0.738 & 0.758 \\
        \hline
        ZeroQ        & 69.36 & 75.56 & 76.89 & 60.17 & 78.99 & 0.681 & 76.01 & 91.59 & 87.90 & 53.34 & \bf{0.739} & 0.718 \\
        ZeroQ-real & 69.29 & 75.57 & 76.89 & \bf{60.18} & 78.87 & 0.681 & 75.87 & \bf{91.62} & \bf{89.29} & 53.29 & 0.738 & 0.675 \\
        Cross-domain & \bf{69.47} & \bf{75.70} & \bf{76.93} & 60.12 & 79.00 & \bf{0.685} & 76.36 & 91.50 & 89.19 & \bf{53.85} & 0.733 & \bf{0.761} \\
        Cross-domain (MS) & \bf{69.47} & \bf{75.70} & \bf{76.93} & 60.08 & \bf{79.01} & 0.684 & \bf{76.37} & 91.50 & 89.19 & 53.38 & 0.732 & \bf{0.761} \\
        \hline
    \end{tabular}
    \caption{8-bit quantization results with different calibration datasets. The dataset abbreviations are defined in Figure~\ref{fig:datasets}. R-18, R-50 and IV3 are Resnet-18, Resnet-50 and InceptionV3 respectively. Cross-domain (MS) uses an average of gram matrices from multiple layers. Best results without using in-domain data are emphasized in bold.}
    \label{tab:compare}
\end{table*}

\begin{figure*}[th]
    \centering
    \begin{subfigure}[t]{0.245\textwidth}
        \centering
        \includegraphics[width=0.99\linewidth]{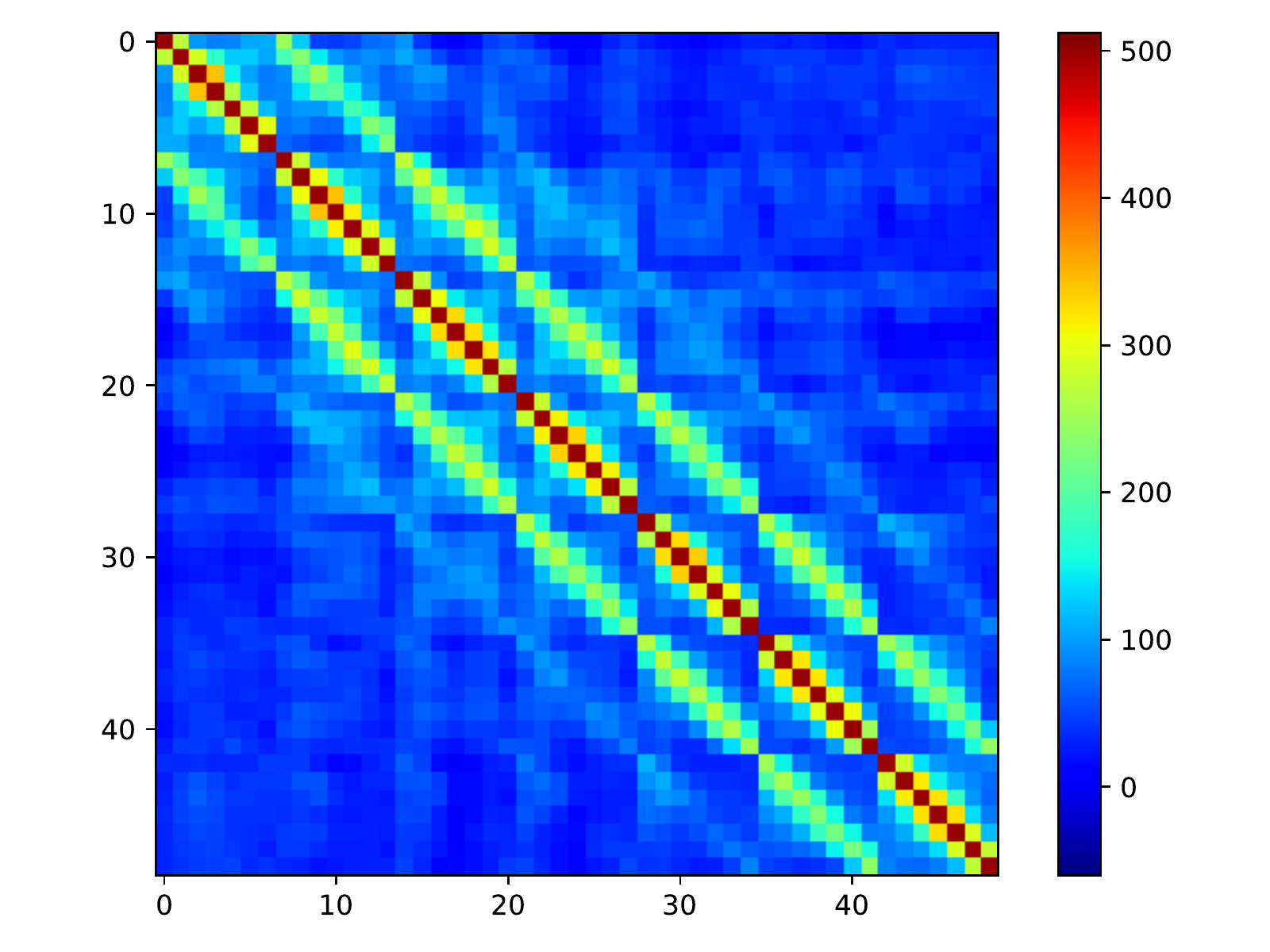}
        \caption{ImageNet}
    \end{subfigure}
    \begin{subfigure}[t]{0.245\textwidth}
        \centering
        \includegraphics[width=0.99\linewidth]{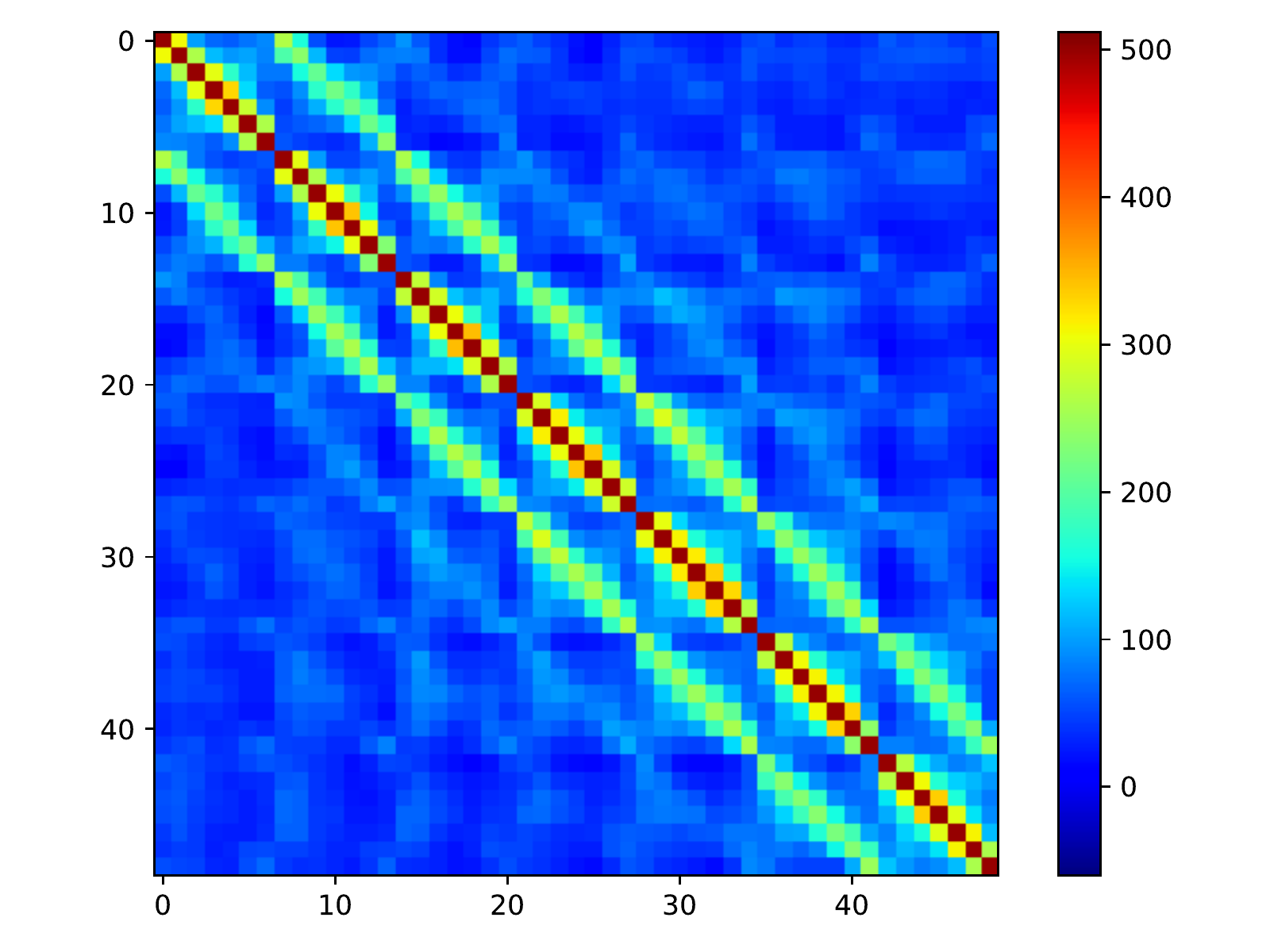}
        \caption{Open Images}
    \end{subfigure}
    \begin{subfigure}[t]{0.245\textwidth}
        \centering
        \includegraphics[width=0.99\linewidth]{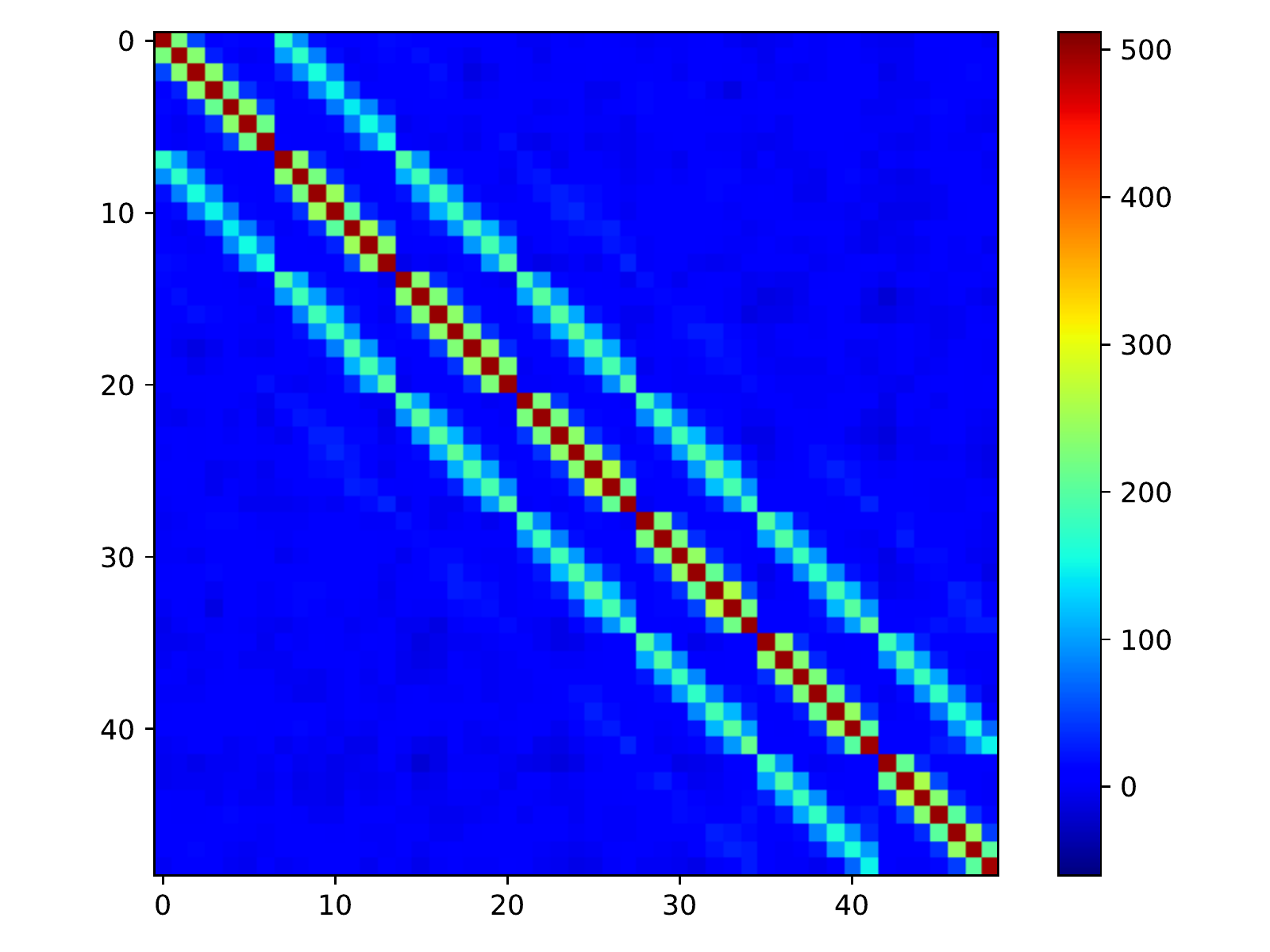}
        \caption{Ultrasound}
    \end{subfigure}
    \begin{subfigure}[t]{0.245\textwidth}
        \centering
        \includegraphics[width=0.99\linewidth]{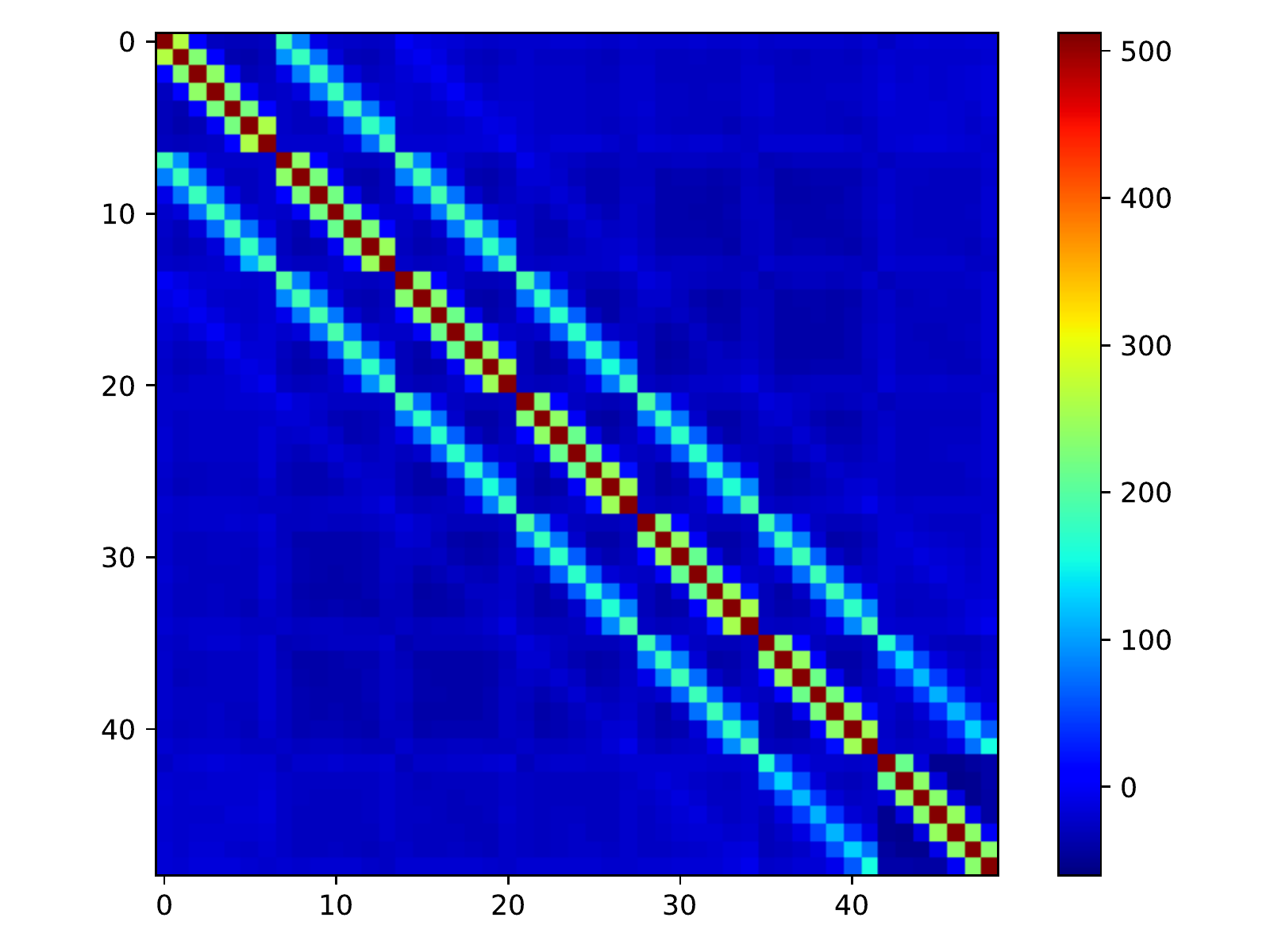}
        \caption{ZeroQ}
    \end{subfigure}
    \caption{Visualization of Gram matrices of different datasets. The feature extraction model is VGG-16 with BatchNorm trained on ImageNet. In Column (b) and (c), BatchNorm adjustment is applied before calculating gram matrices. In Column (d), VGG-16 is used to synthesize the ZeroQ data. Best viewed in color.}
    \label{fig:vis_gram}
    \vspace{-1em}
\end{figure*}

Using the above strategy, we compare our method with in-domain calibration and data synthesis approach ZeroQ~\cite{cai2020zeroq} in Table~\ref{tab:compare}. In addition, we also compare with a variant of ZeroQ that we term as ZeroQ-real, short for ZeroQ with real images. Different from ZeroQ that initializes the synthesized data with random Gaussian values, ZeroQ-real uses real images for initialization. In all the experiments of ZeroQ-real, ImageNet data is used for initialization. We show performance of two variants of our method with one or multiple feature maps in VGG-16 to compute domain discrepancy, shown as Cross-domain and Cross-domain (MS) respectively. When multiple feature maps are used, each feature map will generate a Gram matrix and the domain discrepancy is defined as the average of $L_2$ distances of all Gram matrices. As shown in Table~\ref{tab:compare}, our method achieves comparable or even better results than both ZeroQ and ZeroQ-real. Using one or multiple feature maps to compute domain discrepancy achieves similar performance across different tasks. Our method using selected cross-domain calibration data is on-par with in-domain data, proving the effectiveness of using cross-domain calibration data for post-training quantization.

In Figure~\ref{fig:vis_gram}, we show a visual comparison of gram matrices of different datasets. We show Gram matrices of ImageNet, and two datasets that have the smallest and largest domain discrepancy to ImageNet (\ie, Open Images and Ultrasound). In addition, we also show Gram matrix of synthetic data from ZeroQ. We use VGG-16 with BatchNorm trained on ImageNet as feature extractor. BatchNorm adjustment is applied when using Open Images and Ultrasound datasets. As shown in Figure~\ref{fig:vis_gram}, Open Images shares a similar gram matrix with ImageNet. In contrast, Ultrasound dataset has a very different Gram matrix since it is not natural images. On the other hand, the Gram matrix of the synthetic ZeroQ data is also drastically different from the the two natural image datasets ImageNet and Open Images, showing less spatial correlation than the others. Our hypothesis is that with BatchNorm statistics as the only guidance to synthesize images is not sufficient to build spatial correlations as in real images. The lack of spatial correlation might be a key reason that ZeroQ does not achieve satisfactory performance on tasks such as NIH Chest X-ray classification.

\subsection{Visualization of Activation Ranges}

\begin{figure*}[th]
    \centering
    \includegraphics[width=0.33\linewidth]{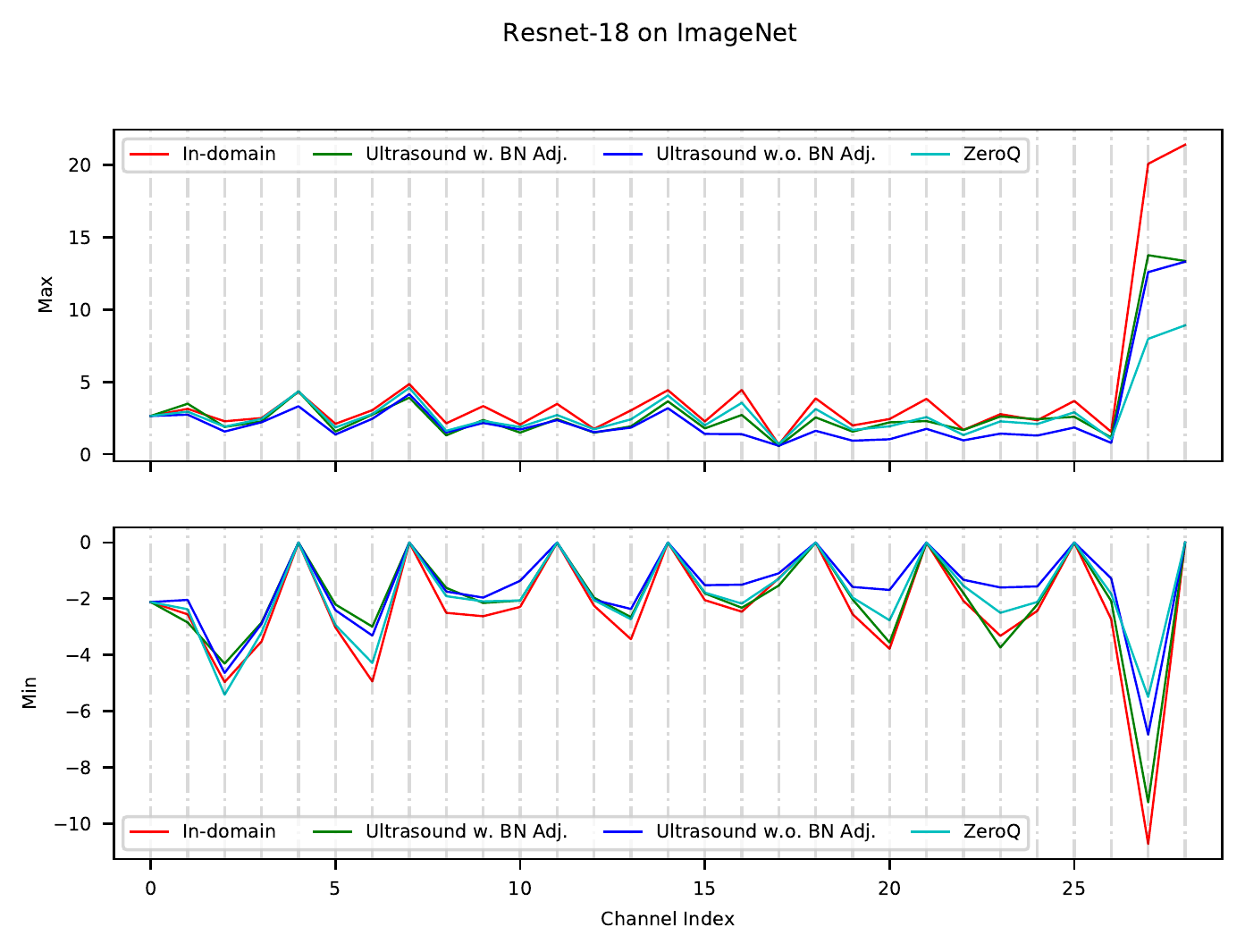}
    \includegraphics[width=0.33\linewidth]{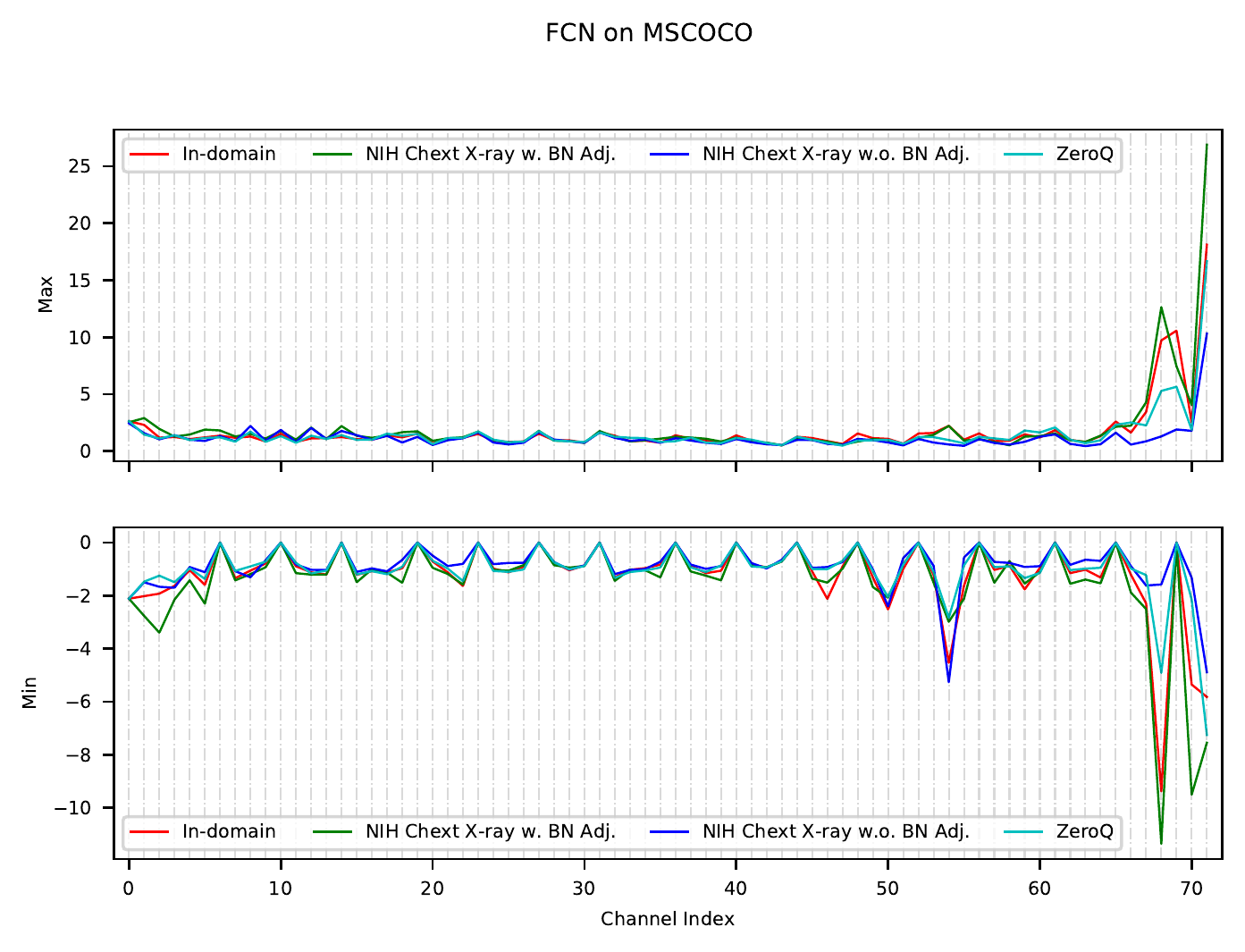}
    \includegraphics[width=0.33\linewidth]{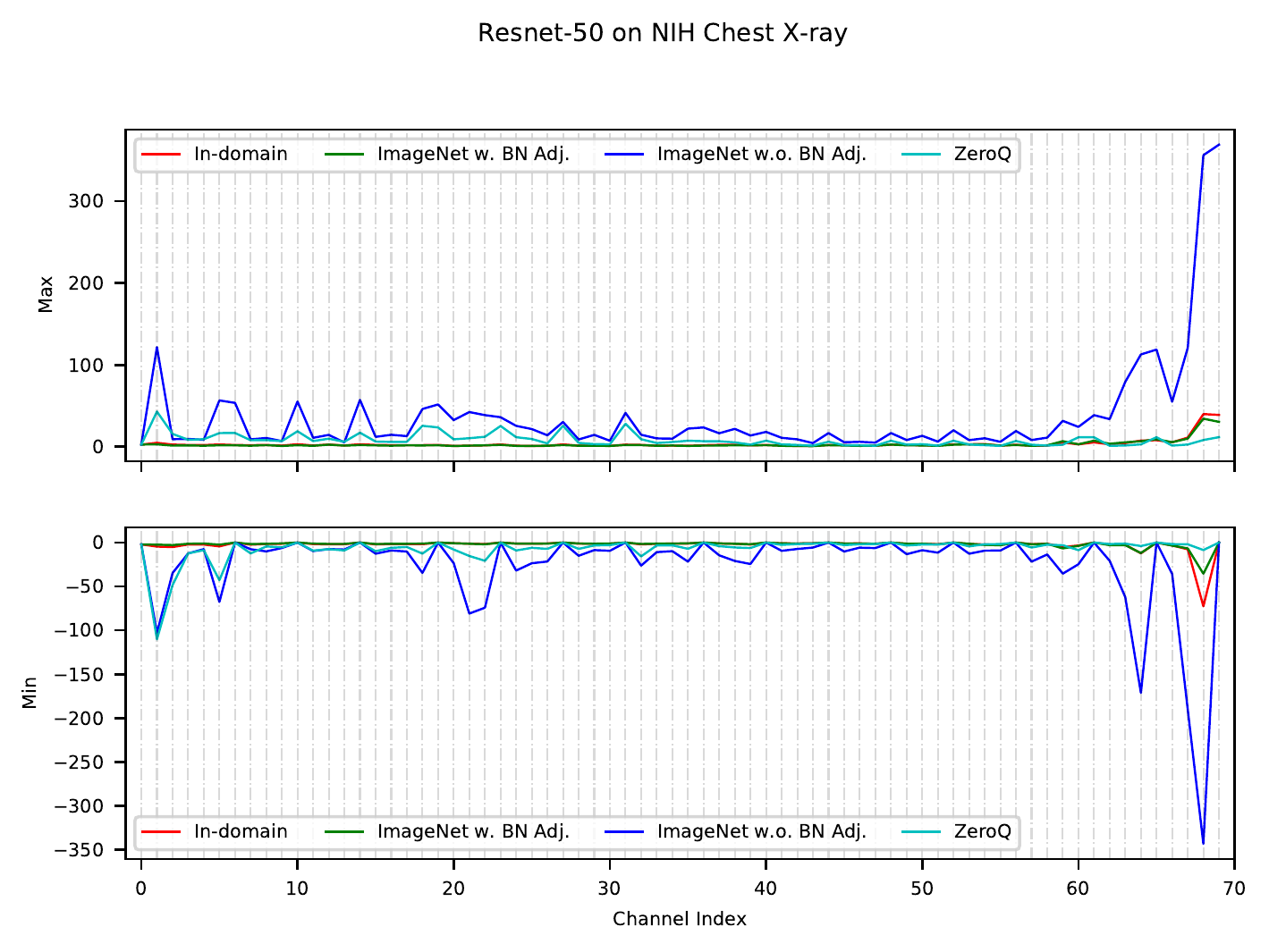}
    \caption{Visualization of calculated activation ranges of different methods. Three examples are shown from left to right: ResNet-18 on ImageNet, FCN on MSCOCO and ResNet-50 on NIH Chest X-ray. In each sub-figure, upper and lower clipping thresholds are plotted separately.}
    \label{fig:act_range}
\end{figure*}

\begin{table*}[th]
    \small
    \centering
    \begin{tabular}{c|C{0.67cm}|C{0.67cm}|C{0.67cm}|C{0.67cm}|C{0.67cm}|C{0.67cm}|C{0.67cm}|C{0.67cm}|C{0.67cm}|C{0.67cm}|C{0.67cm}|C{0.67cm}}
        \hline
        \mr{Calib Methods \textbackslash~Datasets} & \multicolumn{3}{c|}{IN} & \mr{CO} & \mr{CS} & \mr{VO} & \mr{CI} & \mr{CE} & \mr{CB} & \mr{AG}  & \mr{US} & \mr{NI} \\
        \cline{2-4}
         & R18 & R50 & IV3 & & & & & & & &  \\
        \hline
        FP32 & 69.76 & 76.13 & 77.46 & 60.47 & 79.10 & 0.686 & 76.50 & 91.74 & 89.63 & 54.63 & 0.730 & 0.773 \\
        \hline
        In-domain    & 62.19 & 66.80 & 65.81 & 45.97 & 52.57 & 0.612 & 74.86 & 91.00 & 87.01 & 49.11 & 0.739 & 0.711 \\
        \hline
        ZeroQ        & 61.63 & 66.27 & 64.76 & 52.07 & 50.21 & \bf{0.614} & 74.18 & 90.37 & 81.98 & \bf{48.73} & 0.741 & 0.523 \\
        ZeroQ-real & 61.80 & 66.22 & 63.87 & \bf{52.11} & 49.78 & 0.611 & 74.43 & \bf{90.49} & 86.54 & 47.82 & \bf{0.742} & 0.502 \\
        Cross-domain & \bf{62.18} & \bf{66.82} & \bf{65.52} & 50.90 & 53.43 & 0.608 & 74.12 & 90.02 & \bf{87.06} & 48.37 & 0.729 & 0.712 \\
        Cross-domain (MS) & \bf{62.18} & \bf{66.82} & \bf{65.52} & 50.22 & \bf{53.87} & 0.608 & \bf{74.44} & 90.02 & \bf{87.06} & 48.15 & 0.729 & \bf{0.716} \\
        \hline
    \end{tabular}
    \caption{6-bit quantization results with different calibration datasets. The dataset abbreviations are defined in Figure~\ref{fig:datasets}. Best results without using in-domain data are emphasized in bold.}
    \label{tab:compare_6}
    \vspace{-1em}
\end{table*}

In this section, we visualize the activation ranges (clipping thresholds) calculated on different out-of-domain datasets. In Figure~\ref{fig:act_range}, we show three examples: Ultrasound to calibrate ResNet-18 on ImageNet, NIH Chest X-ray to calibrate FCN on MSCOCO, and ImageNet to calibrate ResNet-50 on NIH Chest X-ray. In each sub-figure, we show the lower and upper clipping thresholds for each layer. Compared with the naive cross-domain calibration method, the proposed BatchNorm adjustment makes a closer estimation of activation ranges to those from the in-domain data. Specifically, for ResNet-50 on NIH Chest X-ray, BatchNorm adjustment shows significant improvement over the baseline with original BatchNorm parameters. On the other hand, our estimation is also better than ZeroQ showing that calibration with cross-domain data is more robust than synthetic data in these cases.

\subsection{Lower-bit Quantization}
We also explore cross-domain calibration for 6-bit quantization. We use the same weight and activation quantization schemes as used in 8-bit experiments. Again, we show performance of two variants of our method with one or multiple feature maps to compute domain discrepancy.
The results are summarized in Table~\ref{tab:compare_6}. Best calibration results without using in-domain data are in bold. In most cases, our proposed cross-domain method achieves comparable or better performance than ZeroQ and its variant. Specifically, on Cityscapes and NIH Chest X-ray, our method outperforms ZeroQ by a large margin.

\section{Discussion}
In this work, we show the feasibility of using out-of-domain data for post-training quantization, which is different from the assumption of existing works that in-domain calibration dataset is necessary.
Data synthesis methods are time-consuming and may not perform well in some cases as shown in our experiments. With stable and superior performance on a wide range of tasks, our method can be a new direction of post-training quantization when in-domain data are not available.

There are some interesting topics to explore in future works. First, cross-domain calibration can be improved by designing a better dataset pool consisting of a large amount of diverse domains. Second, better domain discrepancy measures can be explored to further improve the performance of quantized models.

\section{Conclusion}
In this work, we explored cross-domain calibration for post-training quantization. To study this problem, we conducted a large-scale study that spans across various tasks, datasets and neural networks. We find that a simple BatchNorm adjustment strategy can effectively improve the performance of quantized models by a large margin, almost bridging the gap between cross-domain calibration and in-domain calibration. In addition, we find that performance with cross-domain calibration is correlated with Gram matrix similarity between the source and the calibration domains. Therefore, Gram matrix similarity can be used as a criterion to select calibration dataset from a candidate pool to further improve performance. We believe our work will motivate future research on utilizing cross-domain knowledge for network quantization and compression. 

{\small
\bibliographystyle{ieee_fullname}
\bibliography{references}

\begin{thebibliography}{10}\itemsep=-1pt

\bibitem{ultrasound}
Ultrasound nerve segmentation.
\newblock
  \url{https://www.kaggle.com/c/ultrasound-nerve-segmentation/overview}, 2016.

\bibitem{banner2019post}
Ron Banner, Yury Nahshan, and Daniel Soudry.
\newblock Post training 4-bit quantization of convolutional networks for
  rapid-deployment.
\newblock In {\em Advances in Neural Information Processing Systems}, pages
  7950--7958, 2019.

\bibitem{cai2020zeroq}
Yaohui Cai, Zhewei Yao, Zhen Dong, Amir Gholami, Michael~W Mahoney, and Kurt
  Keutzer.
\newblock Zeroq: A novel zero shot quantization framework.
\newblock In {\em Proceedings of the IEEE/CVF Conference on Computer Vision and
  Pattern Recognition}, pages 13169--13178, 2020.

\bibitem{chiu2020agriculture}
Mang~Tik Chiu, Xingqian Xu, Yunchao Wei, Zilong Huang, Alexander~G Schwing,
  Robert Brunner, Hrant Khachatrian, Hovnatan Karapetyan, Ivan Dozier, Greg
  Rose, et~al.
\newblock Agriculture-vision: A large aerial image database for agricultural
  pattern analysis.
\newblock In {\em Proceedings of the IEEE/CVF Conference on Computer Vision and
  Pattern Recognition}, pages 2828--2838, 2020.

\bibitem{cordts2016cityscapes}
Marius Cordts, Mohamed Omran, Sebastian Ramos, Timo Rehfeld, Markus Enzweiler,
  Rodrigo Benenson, Uwe Franke, Stefan Roth, and Bernt Schiele.
\newblock The cityscapes dataset for semantic urban scene understanding.
\newblock In {\em Proceedings of the IEEE conference on computer vision and
  pattern recognition}, pages 3213--3223, 2016.

\bibitem{courbariaux2016binarized}
Matthieu Courbariaux, Itay Hubara, Daniel Soudry, Ran El-Yaniv, and Yoshua
  Bengio.
\newblock Binarized neural networks: Training deep neural networks with weights
  and activations constrained to+ 1 or-1.
\newblock {\em arXiv preprint arXiv:1602.02830}, 2016.

\bibitem{deng2009imagenet}
Jia Deng, Wei Dong, Richard Socher, Li-Jia Li, Kai Li, and Li Fei-Fei.
\newblock Imagenet: A large-scale hierarchical image database.
\newblock In {\em 2009 IEEE conference on computer vision and pattern
  recognition}, pages 248--255. Ieee, 2009.

\bibitem{pascal-voc-2007}
M. Everingham, L. Van~Gool, C.~K.~I. Williams, J. Winn, and A. Zisserman.
\newblock The {PASCAL} {V}isual {O}bject {C}lasses {C}hallenge 2007 {(VOC2007)}
  {R}esults.
\newblock
  http://www.pascal-network.org/challenges/VOC/voc2007/workshop/index.html.

\bibitem{gretton2012kernel}
Arthur Gretton, Karsten~M Borgwardt, Malte~J Rasch, Bernhard Sch{\"o}lkopf, and
  Alexander Smola.
\newblock A kernel two-sample test.
\newblock {\em The Journal of Machine Learning Research}, 13(1):723--773, 2012.

\bibitem{haroush2020knowledge}
Matan Haroush, Itay Hubara, Elad Hoffer, and Daniel Soudry.
\newblock The knowledge within: Methods for data-free model compression.
\newblock In {\em Proceedings of the IEEE/CVF Conference on Computer Vision and
  Pattern Recognition}, pages 8494--8502, 2020.

\bibitem{he2016deep}
Kaiming He, Xiangyu Zhang, Shaoqing Ren, and Jian Sun.
\newblock Deep residual learning for image recognition.
\newblock In {\em Proceedings of the IEEE conference on computer vision and
  pattern recognition}, pages 770--778, 2016.

\bibitem{hubara2016binarized}
Itay Hubara, Matthieu Courbariaux, Daniel Soudry, Ran El-Yaniv, and Yoshua
  Bengio.
\newblock Binarized neural networks.
\newblock In {\em Advances in neural information processing systems}, pages
  4107--4115, 2016.

\bibitem{batchnorm-ioffe15}
Sergey Ioffe and Christian Szegedy.
\newblock Batch normalization: Accelerating deep network training by reducing
  internal covariate shift.
\newblock volume~37 of {\em Proceedings of Machine Learning Research}, pages
  448--456, Lille, France, 07--09 Jul 2015. PMLR.

\bibitem{jin2019efficient}
Qing Jin, Linjie Yang, and Zhenyu Liao.
\newblock Towards efficient training for neural network quantization, 2019.

\bibitem{jung2019learning}
Sangil Jung, Changyong Son, Seohyung Lee, Jinwoo Son, Jae-Joon Han, Youngjun
  Kwak, Sung~Ju Hwang, and Changkyu Choi.
\newblock Learning to quantize deep networks by optimizing quantization
  intervals with task loss.
\newblock In {\em Proceedings of the IEEE Conference on Computer Vision and
  Pattern Recognition}, pages 4350--4359, 2019.

\bibitem{kon2015federated}
Jakub Konečný, Brendan McMahan, and Daniel Ramage.
\newblock Federated optimization:distributed optimization beyond the
  datacenter, 2015.

\bibitem{krishnamoorthi2018quantizing}
Raghuraman Krishnamoorthi.
\newblock Quantizing deep convolutional networks for efficient inference: A
  whitepaper.
\newblock {\em arXiv preprint arXiv:1806.08342}, 2018.

\bibitem{krizhevsky2009learning}
Alex Krizhevsky, Geoffrey Hinton, et~al.
\newblock Learning multiple layers of features from tiny images.
\newblock 2009.

\bibitem{OpenImages}
Alina Kuznetsova, Hassan Rom, Neil Alldrin, Jasper Uijlings, Ivan Krasin, Jordi
  Pont-Tuset, Shahab Kamali, Stefan Popov, Matteo Malloci, Alexander
  Kolesnikov, Tom Duerig, and Vittorio Ferrari.
\newblock The open images dataset v4: Unified image classification, object
  detection, and visual relationship detection at scale.
\newblock {\em IJCV}, 2020.

\bibitem{li2016ternary}
Fengfu Li, Bo Zhang, and Bin Liu.
\newblock Ternary weight networks.
\newblock {\em arXiv preprint arXiv:1605.04711}, 2016.

\bibitem{li2017demystifying}
Yanghao Li, Naiyan Wang, Jiaying Liu, and Xiaodi Hou.
\newblock Demystifying neural style transfer.
\newblock {\em arXiv preprint arXiv:1701.01036}, 2017.

\bibitem{li2016revisiting}
Yanghao Li, Naiyan Wang, Jianping Shi, Jiaying Liu, and Xiaodi Hou.
\newblock Revisiting batch normalization for practical domain adaptation.
\newblock {\em arXiv preprint arXiv:1603.04779}, 2016.

\bibitem{lin2014microsoft}
Tsung-Yi Lin, Michael Maire, Serge Belongie, James Hays, Pietro Perona, Deva
  Ramanan, Piotr Doll{\'a}r, and C~Lawrence Zitnick.
\newblock Microsoft coco: Common objects in context.
\newblock In {\em European conference on computer vision}, pages 740--755.
  Springer, 2014.

\bibitem{Liu_2020_CVPR_Workshops}
Qinghui Liu, Michael~C. Kampffmeyer, Robert Jenssen, and Arnt-Borre Salberg.
\newblock Multi-view self-constructing graph convolutional networks with
  adaptive class weighting loss for semantic segmentation.
\newblock In {\em The IEEE/CVF Conference on Computer Vision and Pattern
  Recognition (CVPR) Workshops}, June 2020.

\bibitem{liu2016ssd}
Wei Liu, Dragomir Anguelov, Dumitru Erhan, Christian Szegedy, Scott Reed,
  Cheng-Yang Fu, and Alexander~C Berg.
\newblock Ssd: Single shot multibox detector.
\newblock In {\em European conference on computer vision}, pages 21--37.
  Springer, 2016.

\bibitem{liu2015faceattributes}
Ziwei Liu, Ping Luo, Xiaogang Wang, and Xiaoou Tang.
\newblock Deep learning face attributes in the wild.
\newblock In {\em Proceedings of International Conference on Computer Vision
  (ICCV)}, December 2015.

\bibitem{liu2018bi}
Zechun Liu, Baoyuan Wu, Wenhan Luo, Xin Yang, Wei Liu, and Kwang-Ting Cheng.
\newblock Bi-real net: Enhancing the performance of 1-bit cnns with improved
  representational capability and advanced training algorithm.
\newblock In {\em Proceedings of the European conference on computer vision
  (ECCV)}, pages 722--737, 2018.

\bibitem{long2015fully}
Jonathan Long, Evan Shelhamer, and Trevor Darrell.
\newblock Fully convolutional networks for semantic segmentation.
\newblock In {\em Proceedings of the IEEE conference on computer vision and
  pattern recognition}, pages 3431--3440, 2015.

\bibitem{nagel2019data}
Markus Nagel, Mart~van Baalen, Tijmen Blankevoort, and Max Welling.
\newblock Data-free quantization through weight equalization and bias
  correction.
\newblock In {\em Proceedings of the IEEE International Conference on Computer
  Vision}, pages 1325--1334, 2019.

\bibitem{netzer2011reading}
Yuval Netzer, Tao Wang, Adam Coates, Alessandro Bissacco, Bo Wu, and Andrew~Y
  Ng.
\newblock Reading digits in natural images with unsupervised feature learning.
\newblock 2011.

\bibitem{nilsback2008automated}
Maria-Elena Nilsback and Andrew Zisserman.
\newblock Automated flower classification over a large number of classes.
\newblock In {\em 2008 Sixth Indian Conference on Computer Vision, Graphics \&
  Image Processing}, pages 722--729. IEEE, 2008.

\bibitem{rastegari2016xnor}
Mohammad Rastegari, Vicente Ordonez, Joseph Redmon, and Ali Farhadi.
\newblock Xnor-net: Imagenet classification using binary convolutional neural
  networks.
\newblock In {\em European conference on computer vision}, pages 525--542.
  Springer, 2016.

\bibitem{sandler2018mobilenetv2}
Mark Sandler, Andrew Howard, Menglong Zhu, Andrey Zhmoginov, and Liang-Chieh
  Chen.
\newblock Mobilenetv2: Inverted residuals and linear bottlenecks.
\newblock In {\em Proceedings of the IEEE conference on computer vision and
  pattern recognition}, pages 4510--4520, 2018.

\bibitem{szegedy2016rethinking}
Christian Szegedy, Vincent Vanhoucke, Sergey Ioffe, Jon Shlens, and Zbigniew
  Wojna.
\newblock Rethinking the inception architecture for computer vision.
\newblock In {\em Proceedings of the IEEE conference on computer vision and
  pattern recognition}, pages 2818--2826, 2016.

\bibitem{WahCUB_200_2011}
C. Wah, S. Branson, P. Welinder, P. Perona, and S. Belongie.
\newblock {The Caltech-UCSD Birds-200-2011 Dataset}.
\newblock Technical Report CNS-TR-2011-001, California Institute of Technology,
  2011.

\bibitem{wang2017chestx}
Xiaosong Wang, Yifan Peng, Le Lu, Zhiyong Lu, Mohammadhadi Bagheri, and
  Ronald~M Summers.
\newblock Chestx-ray8: Hospital-scale chest x-ray database and benchmarks on
  weakly-supervised classification and localization of common thorax diseases.
\newblock In {\em Proceedings of the IEEE conference on computer vision and
  pattern recognition}, pages 2097--2106, 2017.

\bibitem{xu2020generative}
Shoukai Xu, Haokun Li, Bohan Zhuang, Jing Liu, Jiezhang Cao, Chuangrun Liang,
  and Mingkui Tan.
\newblock Generative low-bitwidth data free quantization.
\newblock {\em arXiv preprint arXiv:2003.03603}, 2020.

\bibitem{yu2018bisenet}
Changqian Yu, Jingbo Wang, Chao Peng, Changxin Gao, Gang Yu, and Nong Sang.
\newblock Bisenet: Bilateral segmentation network for real-time semantic
  segmentation.
\newblock In {\em European Conference on Computer Vision}, pages 334--349.
  Springer, 2018.

\bibitem{zhang2018lq}
Dongqing Zhang, Jiaolong Yang, Dongqiangzi Ye, and Gang Hua.
\newblock Lq-nets: Learned quantization for highly accurate and compact deep
  neural networks.
\newblock In {\em Proceedings of the European conference on computer vision
  (ECCV)}, pages 365--382, 2018.

\bibitem{zhang2020threebranch}
Fan Zhang, Meng Li, Guisheng Zhai, and Yizhao Liu.
\newblock Multi-branch and multi-scale attention learning for fine-grained
  visual categorization, 2020.

\bibitem{zhao2019improving}
Ritchie Zhao, Yuwei Hu, Jordan Dotzel, Christopher De~Sa, and Zhiru Zhang.
\newblock Improving neural network quantization without retraining using
  outlier channel splitting.
\newblock {\em arXiv preprint arXiv:1901.09504}, 2019.

\bibitem{zhou2016dorefa}
Shuchang Zhou, Yuxin Wu, Zekun Ni, Xinyu Zhou, He Wen, and Yuheng Zou.
\newblock Dorefa-net: Training low bitwidth convolutional neural networks with
  low bitwidth gradients.
\newblock {\em arXiv preprint arXiv:1606.06160}, 2016.

\bibitem{zhu2016trained}
Chenzhuo Zhu, Song Han, Huizi Mao, and William~J Dally.
\newblock Trained ternary quantization.
\newblock {\em arXiv preprint arXiv:1612.01064}, 2016.

\end{thebibliography}
}

\end{document}